\documentclass[10pt,journal,compsoc]{IEEEtran}
\usepackage{graphicx}
\usepackage{amsmath}
\usepackage{amssymb}
\usepackage{booktabs}
\usepackage{arydshln} 
\usepackage{epsfig}
\usepackage{makecell,rotating}
\usepackage{array}
\usepackage{colortbl}
\usepackage{xcolor}
\usepackage{multirow}
\usepackage{algorithm}  
\usepackage{algpseudocode}

\usepackage[breaklinks,colorlinks]{hyperref}
\usepackage{fancyvrb}
\usepackage{url}

\ifCLASSOPTIONcompsoc
  \usepackage[nocompress]{cite}
\else
  \usepackage{cite}
\fi

\ifCLASSINFOpdf

\else

\fi

\begin{document}

\title{A Survey on Continual Semantic Segmentation: Theory, Challenge, Method and Application}

\author{Bo~Yuan,
	Danpei~Zhao*
\IEEEcompsocitemizethanks{\IEEEcompsocthanksitem Bo Yuan and Danpei Zhao are with the Image Processing Center, School of Astronautics, Beihang University, Beijing 102206, China, and also with the Tianmushan Laboratory, Hangzhou 311115, China.\protect\\
This work was supported by the National Natural Science Foundation of China under Grant 62271018.\protect\\
E-mail: \{yuanbobuaa, zhaodanpei\}@buaa.edu.cn. * Corresponding author.
}
\thanks{Manuscript received XX,XX; revised XX, XX.}}

\markboth{Journal of \LaTeX\ Class Files,~Vol.~X, No.~X, X~X}%
{Shell \MakeLowercase{\textit{et al.}}: Bare Demo of IEEEtran.cls for Computer Society Journals}

\IEEEtitleabstractindextext{%
\begin{abstract}
Continual learning, also known as incremental learning or life-long learning, stands at the forefront of deep learning and AI systems. It breaks through the obstacle of one-way training on close sets and enables continuous adaptive learning on open-set conditions. In the recent decade, continual learning has been explored and applied in multiple fields especially in computer vision covering classification, detection and segmentation tasks. Continual semantic segmentation (CSS), of which the dense prediction peculiarity makes it a challenging, intricate and burgeoning task. In this paper, we present a review of CSS, committing to building a comprehensive survey on problem formulations, primary challenges, universal datasets, neoteric theories and multifarious applications. Concretely, we begin by elucidating the problem definitions and primary challenges. Based on an in-depth investigation of relevant approaches, we sort out and categorize current CSS models into two main branches including \textit{data-replay} and \textit{data-free} sets. In each branch, the corresponding approaches are similarity-based clustered and thoroughly analyzed, following qualitative comparison and quantitative reproductions on relevant datasets. Besides, we also introduce four CSS specialities with diverse application scenarios and development tendencies. Furthermore, we develop a benchmark for CSS encompassing representative references, evaluation results and reproductions, which is available at~\url{https://github.com/YBIO/SurveyCSS}. We hope this survey can serve as a reference-worthy and stimulating contribution to the advancement of the life-long learning field, while also providing valuable perspectives for related fields. 
\end{abstract}

}

\maketitle

\IEEEdisplaynontitleabstractindextext

\IEEEpeerreviewmaketitle

\IEEEraisesectionheading{\section{Introduction}\label{Sec-Introduction}}
\IEEEPARstart{C}{ontinual} learning (CL), which also refers to incremental learning~\cite{wu2019large, castro2018end} or life-long learning~\cite{silver2013lifelong, liu2017lifelong}, is an approach that focuses on acquiring knowledge in a sequential manner. CL originates from cognitive neuroscience research on the mechanisms of memory and forgetting~\cite{Bar2009ThePB, Schrimpf2018BrainScoreWA, vandeVen2020BraininspiredRF, Eryilmaz2014BrainlikeAL} and has experienced prosperous development over the past decade. As a cutting-edge hotspot in deep learning, the CL technique substantially improves the generalization ability of neural network-based models by breaking through the one-off learning constraint. In contrast, conventional machine learning manner normally builds on a close set, i.e., where it can only handle a fixed number of predefined classes, and all the data needs to be presented to the model at the single-step training. However, models often confront the challenge of continuously incremental data in the realm of applicable scenarios. Thus how to enable models to continually adapt to new data or tasks constitutes a prevalent challenge. The primary objective of CL is to strike an optimal balance within the \textit{stability-plasticity dilemma}~\cite{mermillod2013stability} under the constraints of limited computational and storage resources, where stability refers to the capacity to retain previous knowledge and plasticity refers to the ability to integrate new knowledge. 
\begin{figure}[tbp]
	\centering
	\includegraphics[scale=0.45]{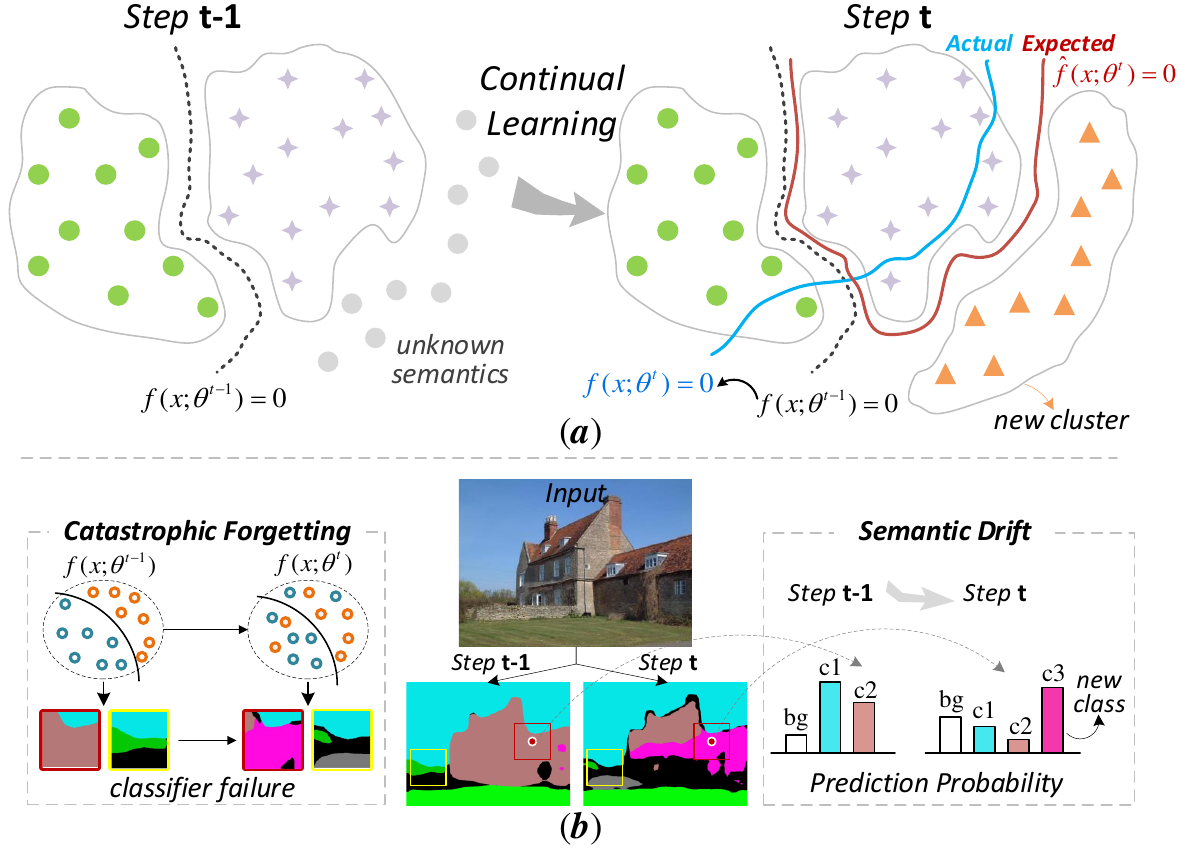}
	\caption{Illustration of catastrophic forgetting and semantic drift in continual semantic segmentation. (a): The decision boundary varies as new data involves, which normally encounters classifier failure. (b): The manifestation of  catastrophic forgetting and semantic drift in CSS, leading to semantic confusion and model degradation reflected in the predicted results.}
	\label{fig-challenge}
\end{figure}

\begin{figure*}
	\centering
	\includegraphics[scale=0.21]{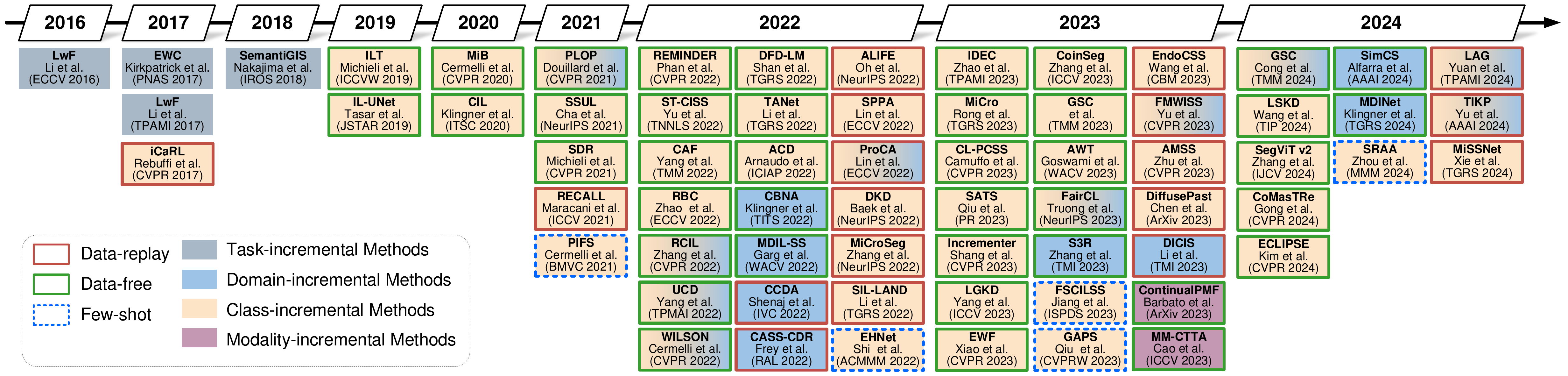}
	\caption{The roadmap of CSS. The representative methods are categorized chronologically. Please note that these methods are not committed to covering all CSS methods but are simply used to validate the taxonomy. Refer to the main text for a more comprehensive summary.}
	\label{fig-roadmap}
\end{figure*}
\begin{figure*}[htbp]
	\centering
	\includegraphics[scale=0.62]{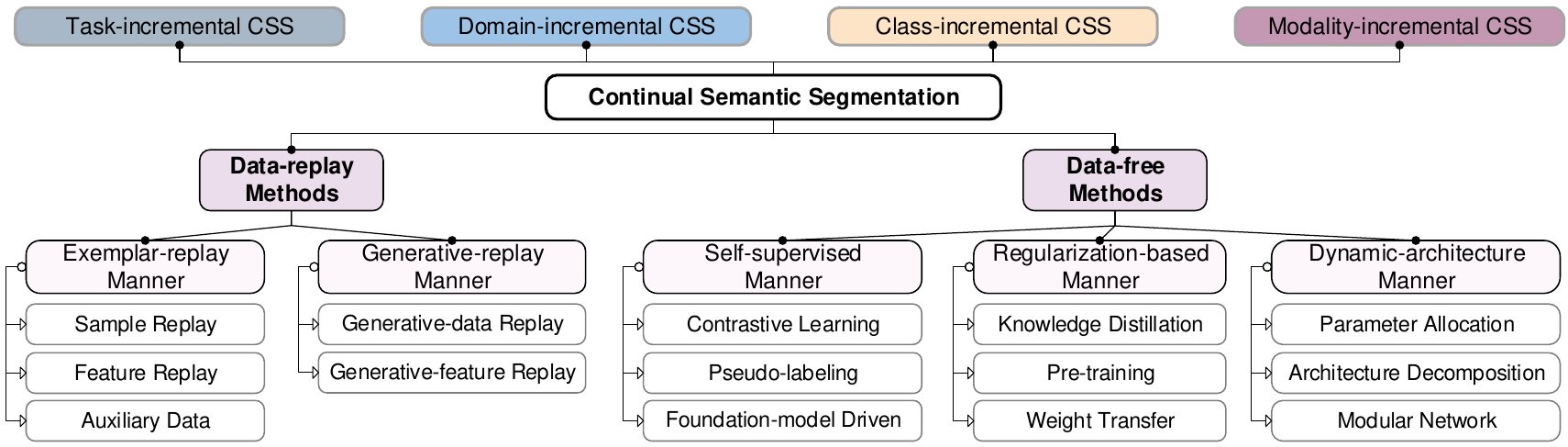}
	\caption{An elaborated taxonomy of continual semantic segmentation methods.}
	\label{fig-method_category}
\end{figure*}

Naturally, the typical model updating involves retraining on new data~\cite{983933} or applying transfer learning techniques~\cite{LWF}, which raises the issue of \textit{catastrophic forgetting}. This problem has been discovered and discussed as early as the 1980s by McCloskey et al.~\cite{McCloskey1989CatastrophicII}. That is algorithms trained with backpropagation suffers from severe knowledge forgetting just like human suffers from gradual forgetting of previously learned tasks.  Additionally, simply re-training the model from scratch can lead to an degradation problem, where the model loses its past ability due to parameter update~\cite{LWF}. As a dense prediction task, continual semantic segmentation (CSS) emerges as a promising but challenging assignment with relevance to various practical vision computing fields such as open-world visual interpretation~\cite{LIU2022OpenworldSS, Wang2022OpenWorldIS}, precision medical assistance~\cite{Zhang2023S3RSA, Ranem2022ContinualHS, bian2021domain}, remote-sensing observation~\cite{Marsocci2022ContinualBT, Feng2021ContinualLW, DAugNet} and autonomous driving~\cite{Vdisch2023CoDEPSOC, ContinualPMF}, etc. 

Besides \textit{catastrophic forgetting}, another critical challenge in CSS is \textit{semantic drift} in the background class at different CL steps. This phenomenon refers to the gradual change or evolution of the semantic content of the background as new classes are incrementally learned. Radically, it roots in the mixed semantics of true background, old classes and future classes. As illustrated in Fig.~\ref{fig-challenge} (a), due to the lack of the historic data, models tend to encounter class confusion and classifier bias during CL steps. In addition, since only the current classes are labeled at each incremental step, the semantics of background pixels undergo a drift because their connotation vary, i.e., known classes and future classes are mixed as the single \emph{background} class. Consequently, it leads to subsequent classification chaos and, ultimately, classifier failures.

As shown in Fig.~\ref{fig-challenge} (b),  the major challenges in CSS encompass catastrophic forgetting and semantic drift. They arise from the absence of old data and parameter updates~\cite{Hu2021DistillingCE, Kaushik2021UnderstandingCF, RW}, leading to semantic confusion and model degradation.  Although a prominent premise in CSS is the inability to access data from old tasks, some research permits the storage of partial old data in a cache to enhance the CSS efficiency when learning new tasks. Additionally, the practical data-free and the eclectic few-shot CSS methods are also currently undergoing in-depth exploration. In Fig.~\ref{fig-roadmap}, we present a chronological list of representative CSS methods, showcasing the evolving research focus over different time periods. It is obvious that the CSS originated and flourished in the recent decade, especially in the last three years. 

Based on the utilization of the historic data, CSS approaches can be broadly categorized into two groups. As depicted in Fig.~\ref{fig-method_category}, the first category, known as \textbf{data-replay} methods, involves storing a portion of past training data as exemplar memory such as~\cite{SSUL, SDR, RECALL, ProCA, DKD, SPPA, ALIFE, FMWISS, AMSS, EndoCSS, DiffusePast}. The second category, termed \textbf{data-free} methods, includes methods like~\cite{ILT, CIL, MiB, PLOP, UCD, REMINDER, RCIL, IDEC, rong2023micro, AWT, EWF, SATS, Incrementer}. These methods utilize transfer learning techniques, such as knowledge distillation (KD), to inherit the capabilities of the old model. Furthermore, there are numerous subcategories of methods, which are summarized in Table~\ref{table-method_analysis} and elaborated in Sec.~\ref{Sec-Methods}. Concerning the application scenarios, CSS methods can also be classified into four kinds of tasks that are detailedly discussed in Sec.~\ref{Sec-Tasks}.
\begin{table*}[htbp]
	\centering
	\caption{Comparison and summary of continual semantic segmentation methods.}
	\setlength{\tabcolsep}{0.2mm}{
		{\begin{tabular*}{0.95\textwidth}{@{\extracolsep{\fill}}ll|ccc@{}}
				\toprule[0.4mm]
				\makecell[c]{Categories} &\makecell[c]{Sub-categories} & Advantages & Disadvantages &Representative\\
				\midrule 
				Exemplar-replay &\makecell[l]{Sample Replay\\Feature Replay\\Auxiliary Data}  &\makecell{strong anti-forgetting, \\easy implementation}&\makecell{storage burdens, \\privacy restrictions} &\cite{SSUL, AMSS, EndoCSS,  kalb2022improving, LAG}\\
				\midrule 
				Generative-replay &\makecell[l]{Generative-data Replay\\ Generative-feature Replay} &\makecell{without storing real data,\\ customized replay}  &\makecell{heavy reliance on \\generative quality,\\high space complexity}  &\cite{RECALL, TIKP, DiffusePast, Liu2022ANG, 9996424}\\
				\midrule 
				Self-supervised &\makecell[l]{Contrastive Learning\\Pseudo-labeling\\Foundation-model Driven} &\makecell{strong adaptability,\\ exemplar-memory free}  &\makecell{high training cost,\\hard to convergence}  &\cite{SDR, UCD, Cermelli2021PrototypebasedIF, ACD, SATS}\\
				\midrule 
				Regularization-based &\makecell[l]{Knowledge Distillation\\Pre-training\\Weight Transfer}&\makecell{quickly updating, \\easy training,\\ low complexity}  &\makecell{classifier shift on new,\\ inefficiency on long-step task}  &\cite{PLOP, IDEC, RCIL, MiB,  ILT}\\
				\midrule 
				Dynamic-architecture&\makecell[l]{Parameter Allocation\\Architecture Decomposition\\Modular Network}&\makecell{high model flexibility,\\strong adaptability to \\diverse data} &\makecell{network parameters \\gradually increases, \\high space complexity} &\cite{ AWT, DKD, liu2020dynamic, yan2021dynamically, ye2022learning}\\
				\bottomrule[0.4mm]
		\end{tabular*}}{}}	
	\label{table-method_analysis}
\end{table*}

Here we would like to discuss the advantage and necessity of continual learning based on specified models during the period of emerging large foundation models. Although recent large-model forms~\cite{SegAnything, wang2023seggpt} achieve fair zero-shot learning ability, they often lack the ability to classify targets with semantic understanding like humans. Another significant concern is cost. For example, large language/vision models usually entail soaring cost for one-time training. And sometimes the historic data becomes inaccessible due to privacy restrictions and storage burdens. Moreover, the need for dedicated models still persists in certain specialized domains such as panoramic remote sensing and medical assistance where high precision is demanded. Therefore, we advocate for the integration of the generality of large models and the customization of specialized models is a future trend. Considering the growing maturity of CL, we believe that this latest and comprehensive survey can provide an overarching perspective for future work. Although there have been some early surveys on continual learning~\cite{Zhou2023DeepCL, Masana2020ClassIncrementalLS, Liu2022IncrementalLW, Wang2023ACS, liu2023incremental, lesort2020continual, belouadah2021comprehensive} with relatively broad coverage, there remains a noticeable gap in reviews that specifically addressing the fundamental dense prediction tasks. Compared to continual learning in image classification~\cite{Wang2023ACS, de2021continual} and object detection tasks~\cite{menezes2023continual}, CSS encounters pixel-wise semantic drift and complex semantic correlation during IL steps, and the dense prediction makes CSS confront more severe forgetting problem. This survey represents a dedicated effort to explore recent advancements in continual semantic segmentation.

The contributions of this paper are outlined as follows.
\begin{itemize}
\item[$\bullet$] This paper reviews the concepts, challenges, methodologies and applications of continual semantic segmentation (CSS), which is a specialized comprehensive survey on this fundamental but flourishing task in the computer vision field. 
\item[$\bullet$] This paper categorizes and summarizes CSS methods based on various technology routes, continual learning strategies and task specifications, which serve as a detailed taxonomy and a comprehensive review of CSS methods.
\item[$\bullet$] We present unified qualitative and quantitative investigations on CSS methods, providing detailed discussions of the advantages, disadvantages and applicable scenarios. 
\item[$\bullet$] We propose an in-depth research analysis on the practical application of CSS and summarize several promising exploration directions.
\end{itemize}

The rest of this paper is organized as follows. Sec.~\ref{Sec-Preliminary} elaborates the basic CSS settings including problem definition, basic formulation and applicable tasks. In Sec.~\ref{Sec-Datasets}, we summarize the datasets and popular protocols of CSS. In Sec.~\ref{Sec-Methods}, up-to-date CSS methods are introduced categorically. Whereafter the qualitative and quantitative analysis and detailed discussions are presented in Sec.~\ref{Sec-Experiments}. Finally, we provide a discussion of current promising applications and summarize the future prospects of CSS in Sec.~\ref{Sec-Applications}. 

\section{Preliminary}
\label{Sec-Preliminary}
\subsection{Problem Definition}
Let $\mathcal{D}=\{(x_i, y_i)\}$ signify the training dataset, where $x_i \in \mathbb{R}^{C\times H\times W}$ denotes the training image and $y_i \in \mathbb{R}^{H\times W}$ denotes the corresponding ground truth. $\mathcal{D}^t$ indicates the training dataset for \emph{t} step. At $t$ step, $C^{0:t-1}$ indicates the previously learned classes and $C^t$ indicates the classes for learning. When training on $\mathcal{D}^t$, the training data of old classes, i.e., $\{\mathcal{D}^0, \mathcal{D}^1, \cdots, \mathcal{D}^{t-1}\}$ is inaccessible. And the ground truth in $\mathcal{D}^t$ only covers $C^t$. The complete training process consists of \{Step-0, Step-1, $\cdots$, Step-T\} steps. Intuitively, models at $t-1$ step and $t$ step are formulated as $M^{t-1}$ and $M^t$.

Considering the infinite persistence of incremental data, at $t$ step, the goal of CSS is to learn a mapping function $f$ parameterized by $\theta$ from the newly added data $\mathcal{D}^{t}=\{(x_i^t,y_i^t)\}_{i=1}^{N^t}$. $f$ aims to minimize the model's loss on $\mathcal{D}^{t}$ while not disrupting the performance of old tasks or data. To achieve this goal, it is crucial to strike a balance between the plasticity of learning new tasks and the stability of maintaining old tasks. Accordingly, the universal objective function for CSS can be defined as:
\begin{equation}
	\min \limits_{\theta^t} \left[ \lambda_1 \mathcal{L}_{base}(\theta^t, \theta^{t-1}, \mathcal{D}^t, C^{0:t-1}) + \lambda_2 \mathcal{L}_{new}(\theta^t, \mathcal{D}^t, C^t)\right]
\end{equation}
where $\mathcal{L}_{new}$ represents the loss functions of new tasks. $\mathcal{L}_{base}$ is to ensure the new model $\theta^t$ to inherit from old model $\theta^{t-1}$. $\lambda_1$ and $\lambda_2$  are coefficients that control the trade-off between old knowledge inheritance and learning of new ones. Of which $\theta^t$ and $\theta^{t-1}$ indicate the model parameter of $t$ step and $t-1$ step, respectively. Specially, it can be formulated as:
\begin{equation}
	\theta^t = \theta^{t-1} - \alpha \nabla \mathcal{L}_t(\theta^{t-1}, \mathcal{D}^t, C^t)
\end{equation}
where $\alpha$ is the learning rate and $\mathcal{L}_{t}$ is the objective function at $t$ step.

\subsection{CSS Tasks}
\label{Sec-Tasks}
\begin{figure}[htbp]
	\centering
	\includegraphics[scale=0.86]{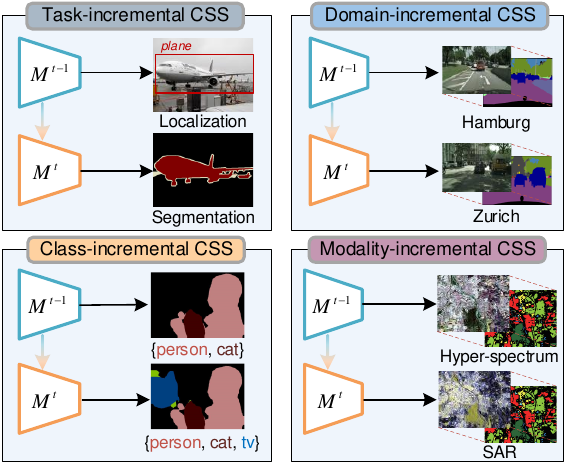}
	\caption{The flowcharts of different CSS specialities.}
	\label{fig-flowcharts}
\end{figure}

In spite of the presentation of an explicit summary of three IL types by~\cite{van2022three}, CSS also encounters various types of tasks. According to the speciality of CL settings, there are mainly four kinds of CSS approaches illustrated in Fig.~\ref{fig-flowcharts}. Concretely,  these specialities encompass:\\
(1) \textbf{Task-incremental CSS}: In this setting, a model is progressively trained to perform new tasks over time. Each new task can involve a different type of prediction or objective, and the model needs to adapt its knowledge while retaining its capability to perform previously learned tasks~\cite{ruvolo2013active, kanakis2020reparameterizing, wallingford2022task, toldo2022learning, roy2023subspace}. \\
(2) \textbf{Domain-incremental CSS}: Domain-incremental learning involves adapting a model to new domains or environments~\cite{tasar2019incremental, 9706918, CCDA, garg2022multi, kalb2021continual, saporta2022multi, michieli2022domain}. This is particularly relevant in cases where a model trained on one dataset needs to generalize to new datasets with different distributions, such as variations in lighting conditions, camera perspectives, or image quality. \\
(3) \textbf{Class-incremental CSS}:  Class-incremental learning emphasizes the gradual incorporation of new classes into a model's inference capacity~\cite{MiB, PLOP, IDEC}. This is a common occurrence in scenarios where the number of classes increases over time, and the model needs to adapt to recognize new classes while preserving its knowledge of previously learned classes.\\
(4) \textbf{Modality-incremental CSS}: Modality-incremental learning deals with incorporating new data modalities into a model's scope. A modality can be a different type of input data, such as adding text data to an existing visual model~\cite{Peng2021AdaptiveMD, zhang2023vqacl, cai2024dynamic, CPP} or introducing data from different sensors~\cite{Li2020MultimodalBF, MM-CTTA}. CSS in this context refers to the model's ability to incorporate and learn from the new modality.

The detailed protocols and objectives of these CSS tasks are also presented in Table~\ref{table-CSS_specialties}. It should be noted that these CSS tasks are not strictly isolated. In many cases, multiple CSS tasks are intertwined such as the class-\&domain-incremental CSS application~\cite{LAG}.	
\begin{table}[htbp]
	\centering
	\caption{The taxonomy of CSS tasks. We categorize CSS into Task-incremental, Domain-incremental, Class-incremental and Modality-incremental scenarios.  It is recommended to analyze this table together with Fig.~\ref{fig-flowcharts}.}
	\setlength{\tabcolsep}{1.0mm}{
		\begin{tabular}{l|c|c}
			\toprule[0.4mm]
			CSS Task&Protocol&Objective \\
			\midrule
			Task-incre.  &\makecell{$\mathcal{D}^{t-1} \cap \mathcal{D}^{t} \neq \emptyset$ \\ $C^{t-1} \cap C^t \neq \emptyset$}&$\arg\min \limits_{M} \sum_{i=1}^t \mathcal{L}(M, \mathcal{D}^i, C^i)$\\
			\midrule
			Domain-incre. &\makecell{$\mathcal{D}^{t-1} \cap \mathcal{D}^{t} = \emptyset$ \\ $C^{t-1}=C^t $}& \makecell{$\arg\min \limits_{M}\sum_{i=1}^{t-1}\mathcal{L}_{base}(M, \mathcal{D}^i)$\\$+\mathcal{L}_{new}(M, \mathcal{D}^t)$}\\
			\midrule
			Class-incre. &\makecell{$\mathcal{D}^{t-1} \cap \mathcal{D}^{t} = \emptyset$ \\ $C^{t-1} \cap C^t =\emptyset$}&\makecell[c]{$\arg\min \limits_{M}\sum_{i=1}^{t-1} \mathcal{L}_{base}(M, \mathcal{D}^i, C^i)$\\ $+\lambda \mathcal{L}_{new}(M, \mathcal{D}^t, C^{t})$}\\
			\midrule
			Modality-incre.  &\makecell{$\mathcal{D}^{t-1} \cap \mathcal{D}^{t} = \emptyset$ \\ $C^{t-1} \cap C^t \neq \emptyset$}&\makecell[c]{$\arg\min \limits_{M}\sum_{i=1}^{t-1} \mathcal{L}_{base}(M, \mathcal{D}^i, C^i)$\\ $+\lambda \mathcal{L}_{new}(M, \mathcal{D}^t, C^{t})$}\\
			\bottomrule[0.4mm]
	\end{tabular}}
	\label{table-CSS_specialties}
\end{table} 

\section{Datasets and Protocols}
\label{Sec-Datasets}
\subsection{Datasets}
\begin{table*}[htbp]
	\centering
	\footnotesize
	\caption{Universal datasets for CSS.}
	\setlength{\tabcolsep}{1.0mm}{
		\begin{tabular}{c|ccccccc}
			\toprule[0.4mm]
			CSS setting&Dataset&Class-num.&Sample-num.&Image size&Format&Content&Year \\
			\midrule
			\multirow{7}*{Domain-incre.}
			&GTA5~\cite{GTA5}&19&24966&1914$\times$1052&RGB&Synthetic urban street scene&2016 \\ 
			&SYNTHIA~\cite{SYNTHIA}&13&9400&1280$\times$760&RGB&Synthetic urban street scene&2016 \\
			&Cityscapes~\cite{Cityscapes}&19&5000&2048$\times$1024&RGB&Urban street scene&2016 \\
			&SemanticKITTI~\cite{SemanticKITTI} &19&23201/20351 scans&4549 points&LiDAR&3D Urban scene&2019 \\
			&ACDC~\cite{ACDC} &19&4006&1920$\times$1080&RGB&Urban street scene&2021\\
			&SHIFT~\cite{sun2022shift}&23&4850 seq.&1280$\times$800&RGB\&LiDAR&Synthetic urban street scene&2022\\
			&SELMA~\cite{testolina2023selma}&19&30909&1280$\times$640&RGB\&LiDAR&Synthetic urban street scene&2022\\
			\midrule
			\multirow{2}*{Class-incre.}&Pascal VOC 2012~\cite{VOC2012}&21&2913&Variable&RGB&wild&2012 \\
			&ADE20K~\cite{ADE}&150&22210&Variable&RGB&indoor\&outdoor&2016\\
			\midrule
			\multirow{5}*{Modality-incre.}&ISPRS-Postdam~\cite{ISPRS}&6&38&6000$\times$6000&RGB-IR&remote-sensing&2013\\
			&ISPRS-Vaihingen~\cite{ISPRS}&6&33&Variable&RG-IR&remote-sensing&2013\\
			&HS-SAR-DSM~\cite{HS-SAR-DSM}&7&78294&332$\times$485&HS-SAR-DSM&remote-sensing&2021\\
			&WHU-OPT-SAR~\cite{WHU-OPT-SAR}&7&100&5556$\times$3704&RGB-SAR&remote-sensing&2022\\
			&FineGrip~\cite{FineGrip} &25 &2649 &800$\times$800 &RGB\&Text&remote-sensing&2024\\
			\bottomrule[0.4mm]
		\end{tabular}
	}
	\label{table-Dataset}
\end{table*} 
Theoretically, any semantic segmentation dataset can be adapted to CSS tasks. Table~\ref{table-Dataset} provides the scenario-specific dataset for CSS tasks. 

Concerning domain-incremental scenarios, CSS models migrate from one domain to another while semantic categories usually keep consistent. For example, Cityscapes~\cite{Cityscapes} consists of 21 urban scenes supporting domain-incremental learning~\cite{PLOP}. ACDC~\cite{ACDC} shares the same classes with Cityscapes but covers four diverse scenario conditions. Considering the requirements of reducing data-annotation dependencies, using synthetic data for training the initial model is a popular way. GTA5~\cite{GTA5} and \mbox{SYNTHIA}~\cite{SYNTHIA} are the representative synthetic datasets that share the common classes with Cityscapes~\cite{Cityscapes}. Some domain-incremental CSS methods~\cite{saporta2022multi, CBNA} have been explored on this benchmark. Recent synthetic datasets~\cite{sun2022shift, testolina2023selma} introduce RGB and LiDAR data for domain-incremental setting, which have the potential to support multi-modal CSS task.

For class-incremental tasks, current CSS methods like~\cite{PLOP, IDEC} separate all classes of the dataset to base classes for initial learning and novel classes for incremental learning. This format allows the model to continuously learn new classes, and the evaluation criteria for this task is the compatibility of both new and old classes. 

For modality-incremental tasks, the model is adapted from one modality to another, which is usually applied in the remote-sensing and cross-modal filed. For example, ISPRS~\cite{ISPRS} provides multiple spectrums for domain- and modality-incremental CSS validation~\cite{LAG}. HS-SAR-DSM~\cite{HS-SAR-DSM} is a multi-modal dataset covering Hyper-Spectrum (HS), Synthetic Aperture Radar (SAR) and Digital Surface Model (DSM). FineGrip~\cite{FineGrip} provides multi-modal data covering captioning and segmentation for remote-sensing panoptic interpretation.

\subsection{CSS Protocols}
According to CSS specialities, the protocols and objectives are summarized in Table~\ref{table-CSS_specialties}.

\textbf{Task-incremental CSS}. It does not strictly limit the inconsistency across datasets and classes. As depicted in Fig.~\ref{fig-flowcharts}, the main concern is to achieve the adaptation and generalization of the model on different tasks. 

\textbf{Domain-incremental CSS}. It requires the overlap between $\mathcal{D}^{t-1}$ and $\mathcal{D}^{t}$ is an empty set but the semantic classes are shared. There are two popular settings including \textit{temporal} and \textit{spatial} CL scenarios. In the \textit{temporal} setting, CSS models need to adapt to changing domains over time to handle variations in the distribution of data at different CL steps. In the \textit{spatial} context, it involves domains across different geographic locations or spatial regions. Thus CSS models need to adapt to semantic segmentation tasks specific to various geographic locations or spatial regions.

\textbf{Class-incremental CSS}. There are two popular class-incremental CSS settings: \emph{disjoint} and \emph{overlapped}. In both settings, only the current classes $C^t$ are labeled and an extra background (bg) class $C^{bg}$. In the former, images at $t$ step only contain $C^{0:t-1} \cup C^t \cup C^{bg}$. While the latter contains $C^{0:t-1} \cup C^{t} \cup C^{t+1: T} \cup C^{bg}$. The disjoint setting uses a unique set of training samples for each training step. Training images in the set depict object/stuff classes belonging to one of the categories to learn in a current step. In the overlapped setting, foreground regions were defined solely within the boundaries of image areas associated with the classes learned during the ongoing stage. Conversely, regions falling outside these bounds, even if they belonged to foreground classes that were previously learned or were scheduled for future learning stages, were classified as background. Similarly, during the testing phase, only those foreground classes that had been learned in the current or earlier stages were considered foreground regions, and all remaining areas were categorized as background. 

\textbf{Modality-incremental CSS}. It requires models continuously adapted new modalities while maintaining the capacity on known knowledge. In this setting, the intersection of $D^t$ and $D^{t-1}$ is an empty set, which is similar to domain-incremental setting. However, the semantic classes are enriched and the modalities vary as the CL steps ongoing. In this setting, CSS models need to overcome the intra-class differences between different modalities and extend the semantic range to multi-modal new data.

\section{Methods}
\label{Sec-Methods}
\begin{figure}[tbp]
	\centering
	\includegraphics[scale=0.9]{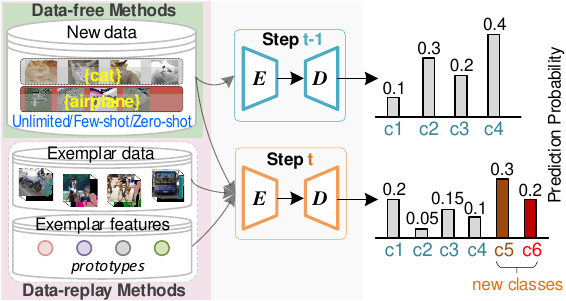}
	\caption{Concerning the dependence on the old data, there are generally data-replay and data-free CSS methods. According to the dependence on the incremental data, data-free branch covers unlimited, few-shot and zero-shot approaches.}
	\label{fig-replay_vs_free}
\end{figure}

In this section, we follow the categorized methods in Fig.~\ref{fig-method_category}, summarizing category-specific representative and up-to-date CSS methods. The generalized processes of data-replay and data-free are depicted in Fig.~\ref{fig-replay_vs_free}. Concretely, data-replay methods are investigated and presented with corresponding illustrations in Sec.~\ref{Sec-Data-Replay}. While data-free approaches are elaborated in Sec.~\ref{Sec-Data-Free}.

\subsection{Data-replay Methods}
\label{Sec-Data-Replay}
An ideal continual learning model does not require storing old data. However, some research proposes to store a small portion of old data as exemplar memory~\cite{SSUL, SATS} or auxiliary data~\cite{RECALL} to assist the model in alleviating catastrophic forgetting. The former combines the old data with new data to participate in model training at CL steps. However, preserving real old data is often constrained in practical applications. On the one hand, as the number of learning tasks increases, the required storage space for preserving old data will become burdensome. On the other hand,  models are not allowed to store training samples in some application domains involving privacy and security concerns. To overcome the aforementioned limitations, generative data-replay methods use a generative model to recover old data. However, such methods are often constrained by the capacity of generative models, and generative models also suffer from forgetting phenomena. In CSS, data-replay methods can be categorized into \textit{exemplar-replay} manner and \textit{generative-replay} manner based on the data-acquiring method. 

\subsubsection{Exemplar-replay Manner}
The main concern of the exemplar-replay manner is to retain the maximum data attributes with a minimum storage cost. It can be divided into \textit{sample-replay}, \textit{feature-replay} and \textit{auxiliary data} methods.

\textit{\textbf{Sample-replay}} methods directly store old images as exemplar memory. As the first sample-replay method in class-incremental learning, iCaRL~\cite{iCaRL} proposes two replay approaches: 1) Fixed total number for all classes. Specifically, assuming the total number of samples is $M$, and the number of learned categories is $C$. The number of stored samples for each class is $m = M/C$. 2) Fixed number for each learned class. In this manner, the storage burden increases as the number of learned classes gradually increases. These two replay manners serve as prototypes for subsequent CSS methods. Following this route, sample selection is also manifold including class-balanced selection, loss-based selection, entropy-based selection, gradient-based selection and representation-based selection, etc. These strategies are analyzed in Table~\ref{table-sample_selection}. 

Current sample-replay methods mainly focus on two aspects. The first is \textbf{\textit{how to select the best samples for replay?}} RECALL~\cite{RECALL}, SSUL-M~\cite{SSUL} and AMSS~\cite{AMSS} propose various sample selection methods to store old data. The future trend in this direction is to store the most representative data to avoid semantic bias. Kalb et al~\cite{kalb2022improving} investigate the influences of various replay strategies for CSS under class- and domain-incremental settings.  The second can be summarized as \textbf{\textit{how to reduce memory storage while retaining the most representative samples?}} Some methods explore small data selection (Kalb et al.~\cite{kalb2022improving}, SSUL-M~\cite{SSUL}) to reduce memory burden. Some methods utilize data augmentation.  Fortin et al.~\cite{fortin2022continual} use copy-paste augmentation to enrich replay semantics. Wang et al.~\cite{EndoCSS} propose a pseudo-replay mechanism within a mini-batch to mitigate storage and privacy issues of exemplar data. And recent work~\cite{TIKP} utilizes text-to-image generation for replaying old data to discard storage burdens.
Concerning domain-incremental CSS, image-style (color~\cite{DICIS}, shape~\cite{Zhang2023S3RSA}, appearance~\cite{chen2023generative, termohlen2021continual}, etc.) are usually considered to inherit the past domain inputs and jointly optimize the new model with incremental data. For instance, Jin et al.~\cite{Kim_2022_CVPR} utilize a meta-learning strategy to build a domain generalization method for semantic segmentation by learning to store domain invariant categorical knowledge in the form of external memory.

\textit{\textbf{Feature-replay}} methods discard the heavy burden of directly storing the original data. Instead, they preserve features or logits and utilize them to optimize the new model, which is more memory-efficient~\cite{chen2024saving}. According to the replay form, this route can be categorized into feature mapping and prototype-alignment approaches. With respect to the former, ALIFE~\cite{ALIFE} propose a feature replay scheme, instead of images directly, to reduce memory requirements. Yoon et al.~\cite{yoon2022semi} adapt a model to the target domain using self-distillation with sample pairs and generate an assistant feature by transferring an intermediate style between the teacher and the student. Yu et. al~\cite{yu2020semantic} propose a metric-learning based embedding network~\cite{wang2021regularizing, chopra2005learning} to preserve known knowledge. 

While prototype-alignment manner preserves old features as prototypes to guide new task learning. Specifically, SDR~\cite{SDR}, PIFS~\cite{Cermelli2021PrototypebasedIF} preserve class-specific prototypes as auxiliary supervision during CL steps.  Lin et al.~\cite{ProCA} utilize prototype alignment for domain-\& class-incremental CSS.  However, the validity of feature prototypes has a crucial impact on the model's continual updating. In other words, insufficient representation capacity of feature prototypes can result in the model lacking discriminative power for features with minimal inter-class differences. On the other hand, when feature prototypes cannot cover the overall data distribution, effective knowledge transfer for data with large intra-class differences is also hindered. In terms of this issue, Shi et al.~\cite{EHNet} propose to use hyper-class knowledge as class-shared semantic properties to enhance the prototype generalization. This enables the new classes to be initialized by a similar known class while focusing on learning discriminative representations, which has been proven effective in few-shot scenarios. Liu et al.~\cite{Liu_2022_CVPR} propose a dynamic prototype convolution network by generating dynamic kernels from a support set, and achieve information interaction using convolution operations over query features. Lin et al.~\cite{SPPA} disentangle the processes of retaining old knowledge and learning new classes, it conducts feature alignment in the encoder and calculates class prototypes in the decoder.  LAG~\cite{LAG} disentangles deep features to semantic-invariant and sample-specific terms for solid prototype preserving. In the remote-sensing field,  Li et al.~\cite{9858901} propose a prototype update mechanism to alleviate the non-adaptive representative prototypes problem. 

Besides directly storing old data or features, introducing \textit{\textbf{auxiliary data}} also benefits alleviating catastrophic forgetting. Such methods often obtain large amounts of unsupervised or weakly supervised data from other areas, such as using a web crawler to draw large amounts of data from the Internet. For example, RECALL-Web~\cite{RECALL} retrieves training examples from online sources. Assuming each learned class tag belonging to $C^{0:t-1}$ can be accessed during $t$-step training, RECALL-Web searches through the website to retrieve images tagged as class $t$ which are fed to the CL training process. Recent large model form achieves very remarkable performance in open-vocabulary tasks. Benefiting from the superior generalization brought by the pre-training on large-scale data, it is possible to reduce the difficulty of model extension on new data. Yu et al.~\cite{FMWISS} utilize a pre-trained foundation model to achieve very competitive CSS performance under weakly-supervised CSS settings. However, the large models normally need fine-tuning to better adapt to specified tasks, which is high-cost in computation resources.
\begin{table}
	\centering
	\caption{Sample selection strategies in exemplar-replay methods.}
	\setlength{\tabcolsep}{0.5mm}{
		\begin{tabular}{l|c|c}
			\toprule[0.4mm]
			Replay Method&Rule&Reference\\
			\midrule
			class-balanced&\makecell{selecting samples that\\ covering every individual class}&\cite{SSUL, yan2021framework}\\
			\hline
			loss-based&\makecell{selecting samples based on\\ the highest, lowest or median\\ value of the cross-entropy loss.} &\cite{9664634}\\
			\hline
			entropy-based&\makecell{the prediction uncertainty\\ is estimated, selecting samples\\ with the lowest, the highest\\ uncertainty, and samples close\\ to the average uncertainty.}&\cite{wiewel2021entropy, EndoCSS, RainbowMC}\\
			\hline
			gradient-based &\makecell{selecting samples based on\\ the diversity of the gradients,\\ keeping the samples with\\ high divergence}&\cite{aljundi2019gradient, GSC, AMSS}\\
			\hline
			representation-based&\makecell{selecting samples based on\\ the distance to the center\\ of all projected samples}&\cite{kalb2022improving}\\
			\bottomrule[0.4mm]
	\end{tabular}}
	\label{table-sample_selection}
\end{table}

\subsubsection{Generative-replay Manner}
In terms of real applications, the exemplar-replay manner is often limited by storage burdens and privacy concerns. While generative replay-based methods introduce generative image replay and generative feature replay.

Previous work has introduced \textit{\textbf{ generative image replay}}, which involves replaying synthetic old class samples generated from a pre-trained GAN~\cite{GAN} or a Diffusion model~\cite{DiffusionModel}. Following this route, RECALL-GAN~\cite{RECALL} retrieves a set of unlabeled replay images for the past semantic classes. However, Chen et al.~\cite{DiffusePast} indicate that GAN-based generative replay suffers from semantic imprecision and encounters out-of-distribution issues, leading to inferior mask annotations and overall performance degradation. Thus they leverage a Stable-Diffusion model~\cite{StableDiffusion} to generate old class images. Thandiackal et al.~\cite{thandiackal2021generative} propose to replay samples that must induce the same hidden features as real samples to train the classifier. In particular, Liu et al~\cite{Liu2022ANG} extend the generative replay approach to medical image semantic segmentation. TIKP~\cite{TIKP} utilizes text-to-image generation for retrieving old data.

With respect to \textit{\textbf{generative feature replay}}, Shan et al.~\cite{9996424} propose to generate pixel-level features for class-incremental CSS in remote-sensing data.

\subsection{Data-free Methods}
\label{Sec-Data-Free}
Data-free methods conduct CSS without storing any old data, aiming to preserve the information about existing classes while making the model progressively learn the new semantics~\cite{Dhar2018LearningWM, 10113287}. It discards the cumbrous memory bank or the additional way to get old data. As seen in Fig.~\ref{fig-method_category}, we categorize the data-free methods to \textit{Self-supervised Manner}, \textit{Regularization-based Manner} and \textit{Dynamic-architecture Manner}.

\subsubsection{Self-supervised Manner}
In the context of CSS, self-supervised learning becomes particularly relevant due to its ability to adapt to new classes or tasks only using labeled incremental data. Self-supervised CSS methods often involve auxiliary tasks like predicting missing pixels, context reconstruction, and image rotations. These tasks guide the model to learn useful features from the available data, enabling it to adapt to new semantics while retaining the knowledge gained from earlier tasks. This direction can be further categorized into three sub-directions. 

\begin{figure}[tbp]
	\centering
	\includegraphics[scale=0.7]{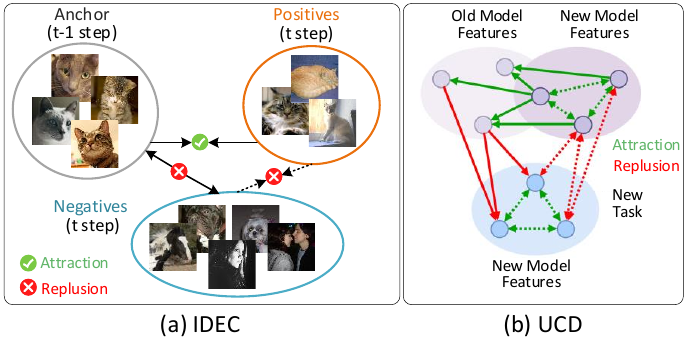}
	\caption{Two typical contrastive learning manners applied in CSS. (a) IDEC~\cite{IDEC}: selecting anchor-class embedding from \textit{t-1} step model, the corresponding positive and negative embeddings from \textit{t} step model. (b) UCD~\cite{UCD}: intra-class attraction and inter-class repulsion between old model features and new model features.}
	\label{fig-contrastive_manner}
\end{figure}

The first kind is \textit{\textbf{contrastive learning}}. The typical paradigm of this manner is introducing proxy tasks with objective functions. For example, contrastive learning can be set in feature or logits alignment~\cite{IDEC, UCD}. With respect to the inner feature distribution, IDEC~\cite{IDEC} proposes a memory-free contrastive learning method named asymmetric region-wise contrastive learning. It extracts reliable anchor embeddings from the old model while positive and negative embeddings from the new model, which is optimized by a triplet loss. Yuan et al.~\cite{LAG} extends the triplet contrastive manner to semantic-invariant features. UCD~\cite{UCD} contrasts features from the new model with features extracted by the previously trained model. We present the depiction of these two typical contrastive learning manners in Fig.~\ref{fig-contrastive_manner}. To reduce the fluctuation during CL, Lin et al.~\cite{lin2023preparing} perform contrastive learning with visual similarity and feature affinity on unseen classes. Zhang et al~\cite{CoinSeg} leverage intra- and inter-class representations to alleviate semantic drift. Besides, metric-learning based methods~\cite{Cen2021DeepML, Dong2022RegionAwareML} are applied in open-world semantic segmentation covering 2D scenes~\cite{frey2022continual, FairCL} to 3D modeling~\cite{Cen2022OpenworldSS, CL-PCSS, riz2023novel, li2023open, kontogianni2024continual, LGKD}. Benefiting by the rich semantic distributions and large intra-class variance, the contrastive learning manner is suitable to be applied in the remote-sensing data~\cite{ACD}.

The second kind is \textit{\textbf{pseudo-labeling}}. This approach utilizes the prediction from the old model as a complement to the supervision for training new model at CL steps. Since the scarcity of labeled data in CSS, it is a popular and effective way to alleviate catastrophic forgetting.  In CSS scenarios, the main striving direction of pseudo-labeling is to avoid the negative optimization problem brought by wrong prediction from the old model to the new model. To achieve this purpose, there are various pseudo-label generation methods have emerged such as class-wise (PLOP~\cite{PLOP}, IDEC~\cite{IDEC}, REMINDER~\cite{REMINDER}) and pixel-wise approaches (ProCA~\cite{ProCA}, ST-CISS~\cite{ST-CISS}, LAG~\cite{LAG}). The former sets different confidence thresholds for different classes. For example, Zhao et al.~\cite{IDEC} propose to set a higher threshold for easy classes while a lower threshold for hard ones to preserve reliable pseudo labels. On the other hand, since the large intra-class variance within dense prediction tasks, some research focuses on measuring pixel-level uncertainty to improve the confidence of pseudo labels~\cite{MagNet, LAG}. 
Recent foundation models are also used to distill the knowledge of complementary foundation models for generating dense pseudo labels~\cite{FMWISS}. 
Considering the high cost of acquiring labeled data, few-shot approaches~\cite{FSCILSS, EHNet, Cermelli2021PrototypebasedIF, dong2018few, tan2024cross} are also explored to reduce the dependence on labeled data. 

The third category is \textit{\textbf{foundation-model driven}}. As a rapid-growing hotspot, foundation models such as the vision-language pre-training (VLP) models~\cite{Jia2021ScalingUV} and the self-supervised pre-training models play an important role in multi-modal research. A representative VLP work is the CLIP series (CLIP~\cite{CLIP}, MaskCLIP~\cite{MaskCLIP}, ZegCLIP~\cite{ZegCLIP}), which jointly trains the image and text encoders on 400 million image-text pairs and achieves zero-shot performance. Recent large-model forms~\cite{SegAnything, Li2023SemanticSAMSA} achieve fair zero-shot learning ability on image segmentation. In CSS, using a strong pre-trained model~\cite{SRAA, oquab2023dinov2} that covers a huge amount of semantic categories can help tackle unseen semantic classes in downstream tasks. Another potential manner is to use prompt learning with foundation models, including visual grounding~\cite{zou2023generalized}, prompt-based segmentation~\cite{luddecke2022image, wang2023seggpt, ECLIPSE}, few-shot personalization incremental segmentation~\cite{zhang2024personalize, liu2023matcher}, etc.
Benefiting from the zero-shot learning and inference ability, the foundation model can be used to drive the weakly-supervised CSS~\cite{FMWISS}, few-shot CSS~\cite{liu2023matcher} and zero-shot CSS~\cite{wang2023seggpt}.

\subsubsection{Regularization-based Manner} 
This direction introduces explicit regularization terms to balance the old and new tasks during CL steps. Depending on the optimization target, the regularization-based manner can be divided into \textit{weight regularization} and \textit{constraint regularization} approaches. Concretely, weight regularization derives task-specific/adaptive parameters~\cite{10214591, ALIFE}. Current CSS approaches usually freeze part of the model's parameters to retain the old capacity. It can effectively limit the sudden drift of neural network weights during CL steps. Constraint regularization normally builds constraint functions on logits or intermediate features between the old and new models. For example, MiB~\cite{MiB}, PLOP~\cite{PLOP}, RBC~\cite{RBC} and IDEC~\cite{IDEC} integrate regular cross-entropy (CE) and knowledge distillation (KD) losses of the background pixels with predictions from the old model. However, the constraint can be built from different patterns. 

\begin{figure}[tbp]
	\centering
	\includegraphics[scale=0.9]{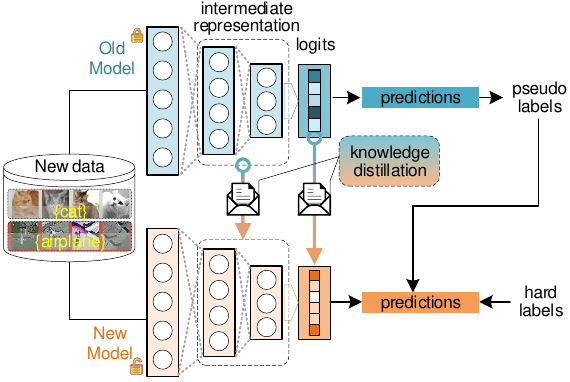}
	\caption{A typical schema of knowledge distillation-based CSS manner.}
	\label{fig-KD}
\end{figure}

The first kind is the \textit{\textbf{knowledge distillation}}. It is a very popular strategy to transfer knowledge from one model (Teacher) to another (Student)~\cite{Yang2019SnapshotDT, Heo2019ACO, Wang2022KnowledgeDA}. KD was firstly defined by~\cite{MC} and generalized by~\cite{Hinton2015DistillingTK}. Considering the dense prediction task, pixel-wise similarity distillation~\cite{Feng2021DoubleSD}, channel-wise distillation~\cite{Shu2021ChannelwiseKD} and layer-wise distillation~\cite{LSKD} are proposed to improve the distillation efficiency. In CSS scenarios, KD has been proven as an effective way to preserve the capability of classifying old classes without storing past data during CL steps.  As seen in Fig.~\ref{fig-KD}, a typical KD-based CSS approach is to use the outputs from the old model (normally the parameters are frozen) to guide the new model (which is trainable) in terms of intermediate representations and logits through customized distillation losses. Following this manner, Michieli et al.~\cite{ILT} explores distillation in intermediate feature space and indicates that L2-norm is superior to cross-entropy or L1-norm. Qiu et al.~\cite{SATS} use self-attention to capture both intra-class and inter-class knowledge. Current methods continually explore the in-depth distillation manners from class weighted~\cite{REMINDER, Incrementer}, objectness guided~\cite{CoMasTRe}, cross-image relationship modeling~\cite{rong2023micro}, prototype rehearsal~\cite{Liu2022IntermediatePM, Cermelli2021PrototypebasedIF, EHNet, SIL-LAND} and cross-scene modeling~\cite{Yang_2023_ICCV}, etc. With respect to the network architecture, some research proves that a stronger backbone is able to improve the distillation performance such as Transformers~\cite{Incrementer, cermelli2023comformer}.  In the remote-sensing field, KD-based CSS has been proven its validity. For instance, Shan et al.~\cite{DFD-LM} perform multi-level feature distillation including both soft distillation and hard distillation on feature representation for class-incremental CSS. MiCro~\cite{rong2023micro} distills the pairwise pixel dependency across mini-batch images in the intermediate feature space. 

The second category is the \textit{\textbf{pre-training}} manner. On the one hand, pre-trained generative models can retrospect old knowledge without storing old data. For instance, Huang et al.~\cite{huang2021half} propose to use a pre-trained image-generative model to invert the trained segmentation network to synthesize input images from random noise. Besides, pre-trained models can be used as an auxiliary task to boost the CSS task. For example, a pre-trained visual saliency model is able to locate regions of interest to further model unknown classes by its intersection with known classes~\cite{SSUL}. Similarly, MicroSeg~\cite{MicroSeg} uses pre-trained Mask2Former~\cite{Mask2Former} as a proposal generator to model unseen classes. On the other hand, recent large models~\cite{SegAnything, Li2023SemanticSAMSA, zhang2024personalize} achieve remarkable performance using large-scale data for pre-training, which shows strong generalization capabilities on multiple tasks like weakly-supervised CSS on only image-level annotations~\cite{WILSON, hsieh2023class}. However, the large model may not always be effective~\cite{Ji2023SegmentAI} without specified constraints for CSS tasks. 
The great potential of large models brings optimistic application prospects for CSS tasks. It still has a long but promising way ahead.

Besides the above two regularization patterns, some methods attempt to utilize \textit{\textbf{weight transfer}} to drive the new model to inherit knowledge from the old model. Zhang et al.~\cite{Zhang2021ComprehensiveIS} build an importance-based selective regularization method for inheritance from the old model. AWT~\cite{AWT} identifies the most relevant weights for new classes from the classifier’s weights for the previous background and transfers these weights to the new classifier. GSC~\cite{GSC} attempts to alleviate the forgetting problem by re-weighting gradient back-propagation for the old classes to optimize the gradient descent. SimCS~\cite{SimCS} uses simulation as a CL regularizer. Summarily, the core objective of the weight transfer manner is to select and transfer the most contributive weight or knowledge from the old model to the new model, to alleviate catastrophic forgetting. 

\subsubsection{Dynamic-architecture Manner}
Many task-incremental CSS methods dynamically extend the network structures during CL steps~\cite{yoon2017lifelong}. For example, Kalb et al.~\cite{10186597} explore the effects of model architectures on CSS tasks. RCIL~\cite{RCIL}  uses a structural re-parameterization mechanism to decouple the representation learning of both old and new knowledge. Klingner et al.~\cite{CBNA} propose a continual unsupervised domain adaptation manner via batchnorm adaptation. According to the model parameter utilization manner, it can be divided into three sub-categories.

The first kind, \textit{\textbf{parameter allocation}} methods allocate a separate parameter space for each incremental task. Concretely, the pioneer LwF~\cite{LWF} models various ways to adapt a model to new tasks. An effective way is to freeze partial parameters to alleviate catastrophic forgetting. Following this protocol, ACD~\cite{ACD} proposes to freeze the old model and utilize it as the teacher to boost the new model updating on new tasks or classes. FairCL~\cite{FairCL} freezes prototypes of old classes to preserve learned knowledge. Moreover, since the model architecture keeps consistent, a solid weight transfer can effectively initialize the new model. On this route,  ALIFE~\cite{ALIFE} and EWF~\cite{EWF} focus on weight transfer and parameter fusion to boost the classifier on new tasks or classes. 

The second way is \textit{\textbf{architecture decomposition}}. This route decomposes the model or parameters into task-specific and task-sharing components. Of which the task-sharing part is able to support reconciling old and new knowledge simultaneously, while the task-specific component is adaptable to incrementally learned tasks. RCIL~\cite{RCIL} proposes a representation compensation using a structural re-parameterization mechanism to boost distillation efficiency. DKD~\cite{DKD} imposes explicit reasoning scores on logits distillation.  Der~\cite{yan2021dynamically} proposes a two-stage learning approach that utilizes a dynamically expandable representation.

The third manner is building \textit{\textbf{modular network}}, which leverages parallel sub-networks or sub-modules to learn incremental tasks in a differentiated manner, without pre-defined task-sharing or task-specific components. Liu et al.~\cite{liu2020dynamic} propose a plug-in module that dynamically constructs and maintains a classifier for the novel class by leveraging the knowledge from the base classes and the information from novel data to overcome the information suppression issue. Ye et al.~\cite{ye2022learning} introduce a concept of flexible knowledge storage and retrieval, where certain knowledge within the network can be temporarily stored in a knowledge bank. When needed, this knowledge can be easily retrieved and reintegrated into the network for operation. This ability for knowledge to be stored and retrieved greatly expands the field of lifelong learning while ensuring user freedom and also serves the purpose of knowledge preservation. 
Yang et al.~\cite{yang2022deep} propose a cordwood-like knowledge transfer strategy that, given a set of pre-trained models trained on different data and heterogeneous architectures, it involves a deep model reassembly process and each model is disassembled into independent model blocks and then these sub-model blocks are selectively reassembled.

\subsection{Other Routes}
Beyond the above exposition, there are some other important creative works in the CSS field.

\textbf{Biological mechanism inspiration}. In CL, biological neural networks often outperform artificial neural networks (ANNs), which impels the investigation of brain-like networks. Caucheteux et al.~\cite{caucheteux2023evidence} map deep language models to brain activity and quantitatively study the similarity between deep language models and the brain when the input content is the same. These results revealed multi-level predictions in the brain.
On the other hand, research on Alzheimer disease~\cite{braak1991neuropathological, knopman2021alzheimer, huang2012alzheimer, blanchard2022dissecting, fouladi2022use} can also help inspire the construction of anti-forgetting measures in CSS. For example, Zhang et al.~\cite{zhang2023brain} propose that in the brain, where an effective and scalable continual learning algorithm appears to have been implemented, the reactivation of neural activity patterns representing previous experiences is believed to be crucial for stabilizing new memories. This memory replay is carefully orchestrated by the hippocampus but is also observed in the cortex, primarily occurring during sharp-wave/ripple events during both sleep and wakefulness. Inspired by this, the authors here reexamine the use of replay as a tool for continual learning in ANNs. Besides,  Refs~\cite{vandeVen2020BraininspiredRF, Wang2020TripleMemoryNA} tackle CL from a brain-inspired manner by bridging the brain activity and ANNs. These studies provide valuable insights for building brain-driven CSS methods.

\textbf{Interdisciplinary study}.  As a cutting-edge research area, CSS is not only rapidly advancing in terms of its theoretical development, but it is also gradually highlighting its significant value in interdisciplinary cross-domain and cross-modality research. Ven et al.~\cite{van2022three} firstly present an explicit summary of three types of incremental learning.  Xu et al.~\cite{xu2024privacy} explore CSS in robotic surgery. Beyond 2D images, there are researches extending CSS to 3D segmentation circumstances~\cite{Yang_2023_CVPR, Yang_2023_ICCV, Liu2022INSConvIS, lin2020active}. These techniques provide vital enlightenment and boosting in autonomous driving. Considering there are sequentially arriving multi-modal data acquired by multi-modal sensors, the joint interpretation for multi-modal incremental data is an urgent task which has been explored from 3D semantic mapping~\cite{li2020incremental, Wu2021SceneGraphFusionI3}, multi-view cooperative interpretation~\cite{chen2019multi, Natan2022EndtoEndAD}, LiDAR data interpretation~\cite{Kang2021ETLiEA},  federated learning~\cite{dong2023federated}, domain generalization~\cite{Muhammad2022VisionBasedSS}, and visual-language collaboration~\cite{Zheng_2023_ICCV}, etc. In the field of remote sensing, research focuses on enhancing small objects~\cite{TANet}, multi-level distillation~\cite{Tasar2019IncrementalLF, DFD-LM, IDEC},  cross-modal distillation~\cite{CPP} and multi-source~\cite{DAugNet} unsupervised domain-incremental CSS.

\begin{table*}[htbp]
	\caption{Qualitative comparison of CSS methods. Rating system follows:  If the model's performance exceeds 25\%, 50\% and 75\% of the offline setting, one, two and three $\bigstar$ are marked, respectively. }
	\centering
	\scriptsize
	\setlength{\tabcolsep}{0.5mm}{
		{\begin{tabular*}{0.95\textwidth}{@{\extracolsep{\fill}}l|ccccccc@{}}
				\toprule[0.5mm]
				Method &\makecell{Published\\Year}&\makecell{Replay\\Based}&Purpose& \makecell{Testing\\Benchmark}&\makecell{Anti-forgetting\\on Old}&\makecell{Accuracy\\ on New}&\makecell{Code\\Available}\\
				\midrule
				EWC~\cite{EWC}                           &PNAS 2017        &-                  &Task-incre.                     &MNIST		               &$\star$                  &$\star$                  &-\\
				iCaRL~\cite{iCaRL}                          &CVPR 2017        &-                 &Task-incre.                     &CIFAR-100\&ILSVRC&$\star$                  &$\star$                  &-\\
				LwF~\cite{LWF}                              &TPAMI 2017       &-                  &Task-\&Domain-incre.    &ImageNet\&Places365\&VOC2012&$\star$                   &$\star$                  &$\checkmark$ \\
				ILT~\cite{ILT}                                  &ICCVW 2019       &-                 &Class-incre.                  &VOC2012			            & $\bigstar$                   &$\bigstar$  				 &$\checkmark$ \\
				MiB~\cite{MiB}                                &CVPR 2020        &-                  &Class-incre.                  &VOC2012\&ADE20K         &$\bigstar\bigstar$                  & $\bigstar$               &$\checkmark$ \\
				PLOP~\cite{PLOP}                          &CVPR 2021        &-                  &Class-\&Domain-incre.  &VOC2012\&ADE20K\&Cityscapes 		 & $\bigstar\bigstar\bigstar$                 &$\bigstar$          &$\checkmark$\\
				SDR~\cite{SDR}                             &CVPR 2021        &-                    &Class-incre.                 &VOC2012\&ADE20K 		   &$\bigstar\bigstar$           &$\bigstar$           &$\checkmark$ \\
				RECALL~\cite{RECALL}                 &ICCV 2021         &$\checkmark$&Class-incre.                 &VOC2012						  &$\bigstar\bigstar$            & $\bigstar\bigstar$           &$\checkmark$ \\
				SSUL~\cite{SSUL}                         &NeurIPS 2021     &$\checkmark$&Class-incre.                  &VOC2012\&ADE20K	      &$\bigstar\bigstar\bigstar$ 	  &$\bigstar\bigstar$    &$\checkmark$ \\
				UCD~\cite{UCD}                            &TPAMI 2022       &-                    &Class-\&Domain-incre.  &	VOC2012\&ADE20K\&Cityscapes  &$\bigstar\bigstar$ 			&$\bigstar$         &$\checkmark$ \\
				CAF~\cite{CAF}                             &TMM 2022          &-                   &Class-incre.                  &VOC2012\&ADE20K			 &$\bigstar\bigstar\bigstar$		     &$\bigstar\bigstar$			 &$\checkmark$ \\
				RCIL~\cite{RCIL}                           &CVPR 2022         &-                    &Class-incre.                  &VOC2012\&ADE20K			&$\bigstar\bigstar\bigstar$		&$\bigstar$              &$\checkmark$ \\
				REMINDER~\cite{REMINDER}        &CVPR 2022        &-                    &Class-incre.                 &VOC2012\&ADE20K         &$\bigstar\bigstar\bigstar$ 		      &$\bigstar\bigstar$           &- \\
				WILSON~\cite{WILSON}               &CVPR 2022       &-                      &Class-\&Domain-incre. &VOC2012\&COCO				&$\bigstar\bigstar$       &$\bigstar\bigstar$       &$\checkmark$ \\
				ST-CISS~\cite{ST-CISS}                &TNNLS 2022       &-                    &Class-incre.                 &VOC2012\&ADE20K	   	  & $\bigstar\bigstar\bigstar$		  & $\bigstar\bigstar$           &$\checkmark$ \\
				CBNA~\cite{CBNA}                       &TITS 2022           &-                   &Domain-incre.             &GTA5\&SYNTHIA\&Cityscapes\&KITTI		 &$\bigstar\bigstar$ &$\bigstar\bigstar$  &$\checkmark$ \\
				MicroSeg~\cite{MicroSeg}           &NeurIPS 2022     &$\checkmark$&Class-incre.                  &VOC2012\&ADE20K	      &$\bigstar\bigstar\bigstar$ 	  &$\bigstar\bigstar\bigstar$    &$\checkmark$ \\
				DKD~\cite{DKD}                           &NeurIPS 2022     &$\checkmark$ & Class-incre.                &VOC2012\&ADE20K			 &$\bigstar\bigstar\bigstar$   &$\bigstar\bigstar$     &$\checkmark$ \\
				ALIFE~\cite{ALIFE}                     &NeurIPS 2022     &$\checkmark$ & Class-incre.                &VOC2012\&ADE20K			 &$\bigstar\bigstar\bigstar$  &$\bigstar\bigstar\bigstar$ 	&$\checkmark$ \\
				SPPA~\cite{SPPA}                        &ECCV 2022        &$\checkmark$ &Class-incre.                 &VOC2012\&ADE20K	   	      &$\bigstar\bigstar\bigstar$  	    &$\bigstar$ 		   &$\checkmark$ \\
				RBC~\cite{RBC}                            &ECCV 2022       &-                     &Class-incre.                &VOC2012\&ADE20K						 &$\bigstar\bigstar\bigstar$				 &$\bigstar\bigstar$			  &$\checkmark$ \\
				IDEC~\cite{IDEC}                        &TPAMI 2023        &-                     &Class-incre.                &VOC2012\&ADE20K\&ISPRS &$\bigstar\bigstar\bigstar$  &$\bigstar\bigstar$          &$\checkmark$ \\
				MiCro~\cite{rong2023micro}         &TGRS 2023         &-                      &Class-incre.                 &ISPRS\&iSAID              &$\bigstar\bigstar$         &$\bigstar\bigstar$        &$\checkmark$ \\
				FairCL~\cite{FairCL}                  &NeurIPS 2023      &-                         &Class-\&Domain-incre.   &VOC2012\&ADE20K\&Cityscapes  &$\bigstar\bigstar\bigstar$		&$\bigstar\bigstar$              &-\\
				FMWISS~\cite{FMWISS}          &CVPR 2023        &$\checkmark$ &Class-\&Domain-incre. &VOC2012\&COCO    			  &$\bigstar\bigstar\bigstar$  &$\bigstar\bigstar$ 		 &- \\
				EWF~\cite{EWF}                         &CVPR 2023        &-                      &Class-incre.               &VOC2012\&ADE20K              &$\bigstar\bigstar\bigstar$		   &$\bigstar$		&- \\
				Incrementer~\cite{Incrementer}     &CVPR 2023        &-                      &Class-incre.	             &VOC2012\&ADE20K               &$\bigstar\bigstar\bigstar$  &$\bigstar\bigstar\bigstar$   &-\\
				AMSS~\cite{AMSS}                      &CVPR 2023        &$\checkmark$  &Class-incre.	            &VOC2012\&ADE20K              &$\bigstar\bigstar\bigstar$	&$\bigstar\bigstar$          &- \\	
				AWT~\cite{AWT}                         &WACV 2023       &-                       &Class-incre.               &VOC2012\&ADE20K  		        &$\bigstar\bigstar\bigstar$	     &$\bigstar\bigstar$		   &$\checkmark$ \\
				SATS~\cite{SATS}                        &PR 2023            &-                      &Class-incre.               &VOC2012\&ADE20K               &$\bigstar\bigstar\bigstar$	   &$\bigstar\bigstar\bigstar$  &$\checkmark$ \\
				GSC~\cite{GSC}                        & TMM 2024		    &-                        &Class-\&Domain-incre. &VOC2012\&ADE20K\&Cityscapes &$\bigstar\bigstar\bigstar$		&$\bigstar\bigstar$              &$\checkmark$ \\
				TIKP~\cite{TIKP}                        &AAAI 2024            &$\checkmark$  &Class-\&Domain-incre.   &VOC2012\&ADE20K\&Cityscapes &$\bigstar\bigstar\bigstar$		&$\bigstar\bigstar$              &-  \\
				SimCS~\cite{SimCS}               &AAAI 2024           &-                       &Domain-incre.                     &Cityscapes\&IDD\&BDD\&ACDC    &$\bigstar\bigstar\bigstar$		&$\bigstar\bigstar$             &-\\
				ECLIPSE~\cite{ECLIPSE}       &CVPR 2024        &-                         &Class-incre.                       &ADE20K\&COCO                            &$\bigstar\bigstar\bigstar$		&$\bigstar\bigstar$             &$\checkmark$\\
				CPP~\cite{CPP}               &ACMMM 2024        &-  &Class-incre. &FineGrip &$\bigstar\bigstar$  &$\bigstar\bigstar$          &$\checkmark$ \\
				LAG~\cite{LAG}                           &TPAMI 2024        &$\checkmark$  &Class-\&Domain-incre. &VOC2012\&ADE20K\&ISPRS &$\bigstar\bigstar\bigstar$  &$\bigstar\bigstar$          &$\checkmark$ \\
				\bottomrule[0.5mm]
		\end{tabular*}}{}}	
	\label{table-QC}
\end{table*}

\section{Performance Evaluation and Analysis}
\label{Sec-Experiments}
\subsection{Evaluation Metrics}
The evaluation of CSS tasks mainly encompasses two aspects: \textit{accuracy} and \textit{forgetfulness}. Of which the accuracy measures the testing precision of all learned tasks after all CL steps, while the forgetfulness gauges the extent of the average performance drop after all CL steps. Typically, the accuracy is defined as :
\begin{equation}
	A_t = \frac{1}{t} \sum_{i=1}^{t} a_{i}
\end{equation}
where $A_t$ represents the model's performance on all seen tasks $C^{0:t}$ at step $t$.  $a_{i}$ indicates the accuracy at $i$ step. 

The forgetfulness is calculated by:
\begin{equation}
	F_t = \frac{1}{t} \sum_{i=1}^{t} (\frac{|a_{0}-a_{i}|}{a_{0}}) 
\end{equation}
where $F_t$ is the average forgetfulness at $t$ step. $a_0$ is the accuracy at the initial learning step while $a_i$ indicates the accuracy at $i$ step. 

Recently there are some research direct at CL evaluation. The index of CL score proposed in~\cite{diaz2018don} is defined as
\begin{equation}
	CL_{score} = \sum_{i=1}^{\mathcal{C}} w_i c_i 
\end{equation}
where $c_i\in \mathcal{C}, c_i\in [0,1]$ represents each criterion belonging to all criterions $\mathcal{C}$ and weight $w_i\in [0,1]$ satisfies $\sum_{i=1}^\mathcal{C} w_i=1$. 
Mirzadeh et al.~\cite{mirzadeh2022architecture} concern the evaluation in CL via four aspects including average accuracy, learning accuracy, joint accuracy and average forgetting, which also cover learning ability and forgetting measurement.

For dense prediction task, the popular metric is the mean intersection over union (mIoU), which is calculated by:
\begin{equation}
	IoU = \frac{TP}{TP+FP+FN}
\end{equation}
where TP, FP and FN are the numbers of true positive, false positive and false negative pixels, respectively. Specifically, in CSS tasks, it is common to simultaneously report the mIoU on old, new and all average tasks or domains or classes. Another is the Dice metric, which is formulated as:
\begin{equation}
	Dice = \frac{2 \times TP}{TP+2 \times FP+FN}
\end{equation}

\subsection{Qualitative Comparison}
\label{Sec-Exp_QC}
We compare current CSS methods in terms of publication date, old-data dependence, purpose, testing benchmark, anti-forgetting performance on old and accuracy on new tasks in Table~\ref{table-QC}. 

Data-free methods address catastrophic forgetting and classifier failure problems without old data inference. As seen in Table~\ref{table-QC}, ILT~\cite{ILT}, MiB~\cite{MiB}, PLOP~\cite{PLOP}, DFD-LM~\cite{DFD-LM} utilize multi-level knowledge distillation covering intermediate representations and output logits. RCIL~\cite{RCIL} and DKD~\cite{DKD} emphasize the significance of addressing semantic drift, particularly in CSS. Following this route, IDEC~\cite{IDEC}, UCD~\cite{UCD} and ACD~\cite{ACD} introduce contrastive learning to CSS to mitigate semantic drift between old and new classes. However, most CSS methods typically build upon an existing semantic segmentation method, such as DeepLabv3~\cite{DeepLabv3}, which raises a question that \textbf{\textit{Does the semantic segmentation model itself affect CSS performance?}} To address this issue, Kalb et al.~\cite{10186597} study how the choice of neural network architecture affects catastrophic forgetting in class- and domain-incremental CSS tasks. Earlier Yuan et al.~\cite{BAFFT} discuss the impact of various semantic models and backbones on domain-incremental CSS. It proposes a novel metric namely Normalized Adaptability Measure (NAM) to evaluate the improvement of CSS performance. Zhao et al.~\cite{IDEC} and Yuan et al~\cite{LAG} investigate the CSS performance by using CNN and Transformer architectures. Refs.~\cite{Incrementer, zhang2024segvit} utilize ViT~\cite{ViT} to achieve favourable performance. The above researches demonstrate that a stronger semantic segmentation model can help achieve superior CSS performance. However, of course, the dataset distribution and application scenarios are also vital factors in determining CSS performance.

Replay-based methods leverage old data or semantics for explicitly retrieving old knowledge. Such methods usually achieve favourable anti-forgetting ability on old tasks or classes such as SSUL-M~\cite{SSUL}, DKD~\cite{DKD}, etc. However, the various replaying strategies are effected by specific samples and the order of the training samples, it may limit the generalization of the model when encountering large semantic gap between old and new tasks. Feature-replay~\cite{LAG} and generative-replay~\cite{TIKP} methods reduce the storage burdens but also maintains favourable performance.

Besides minimizing the old data dependence, the optimization on reducing reliance on the labeled incremental data is a burgeoning direction in CSS.  EHNet~\cite{EHNet}, FSCILSS~\cite{FSCILSS} and SRAA~\cite{SRAA} introduce few-shot settings to CSS. The main challenges of few-shot CSS lie in feature drift on old classes and overfitting issues on new classes. Thus hyper-class representation embedding~\cite{EHNet}, cross-image relationship modeling~\cite{PANet} and pseudo-labeling\cite{FSCILSS} are normally used to boost the performance. Exploiting unlabeled images as auxiliary data is also a promising way. Another interesting and effective manner is the foundation-model driven method. For example, FMWISS~\cite{FMWISS} uses pre-training-based co-segmentation to distill the knowledge of complementary foundation models. It resorts to the strong zero-shot learning ability of large models to achieve weakly-supervised CSS by generating dense pseudo labels from image-level labels. With the rapid growth of large models, we believe the CSS problem will encounter a promising in-depth study.

\subsection{Quantitative Analysis}
In this section, we report the quantitative results of the representative up-to-date CSS models. Concretely, we evaluate the CSS methods under class-incremental and domain-incremental CSS settings, respectively.  

\subsubsection{Class-incremental CSS Evaluation}
Aligning with the categorization in Sec.~\ref{Sec-Methods}, we provide the quantitative results for data-free and date-replay manners, respectively. To comprehensively evaluate the anti-forgetting and adapting performance of the models, we organize it in three ways: few-step with multi-class (FSMC), multi-step with few-class (MSFC), multi-step with multi-class (MSMC). Particularly, FSMC emphasizes the ability to learn new knowledge (\textbf{plasticity}) since many new classes/tasks are adapted in a single step. In contrast, MSFC underlines the ability of anti-forgetting on old knowledge (\textbf{stability}) because many CL steps are conducted. MSMC synchronously measures the ability of anti-forgetting and learning new knowledge. The quantitative investigations are conducted on Pascal VOC 2012~\cite{VOC2012} and ADE20K~\cite{ADE}.

\noindent\textbf{Pascal VOC 2012}. On Pascal VOC 2012, we evaluate the CSS models on 15-5 (2 steps), 15-1 (6 steps), 5-3 (6 steps) and 10-1 (11 steps) settings. For example, 
15-1 indicates initially learning 15 classes and then learning the additional one class at each step for a total of another 5 steps. Of which VOC 15-5 can be considered as FSMC setting,  VOC 15-1 and VOC 10-1 are MSFC manners while VOC 5-3 is deemed as MSMC setting. The results are conducted on the \textit{disjoint} and the \textit{overlapped} CSS settings with a greater focus on the latter due to its realistic peculiarity.

\begin{table*}[htbp]
	\caption{Class-incremental CSS quantitative comparison on Pascal VOC 2012 in mIoU (\%) under \textit{disjoint} abd \textit{overlapped} settings. Class 0 indicates the unlabeled class.  Methods with * indicate the results were directly taken from the corresponding original work, and all the others were based on our re-implementation.}
	\centering
	\scriptsize
	\setlength{\tabcolsep}{0.5mm}{
		{\begin{tabular*}{0.95\textwidth}{@{\extracolsep{\fill}}ll|c|c|ccc|ccc|ccc|ccc@{}}
				\toprule[0.5mm]
				&\multirow{2}*{Method} &\multirow{2}*{Year}&\multirow{2}*{Model}&
				\multicolumn{3}{c}{15-5 (2 steps)} & \multicolumn{3}{c}{15-1 (6 steps)} & \multicolumn{3}{c}{5-3 (6 steps)} & \multicolumn{3}{c}{10-1 (11 steps)} \\
				&&&& 0-15 & 16-20 & all & 0-15 & 16-20 & all & 0-5 & 6-20 & all & 0-10 & 11-20 & all\\
				\midrule
				\multicolumn{15}{c}{Disjoint} \\
				\midrule
				&\emph{fine tuning}&-&DeepLabv3&1.10&33.60&9.20&0.20& 1.80& 0.60&2.10&1.30&1.50&6.30&1.10&3.80\\
				&MiB*~\cite{MiB}&CVPR2020&DeepLabv3 &71.80&43.30&64.70 &46.20&12.90&37.90&-&-&-&9.50&4.10&6.90 \\
				&PLOP*~\cite{PLOP}&CVPR2021&DeepLabv3&71.00&42.82&64.29 &57.86&13.67&46.48&-&-&-&9.70&7.00&8.40 \\
				&SDR~\cite{SDR} &CVPR2021&DeepLabv3+ &74.60&44.10&67.30 &59.40&14.30&48.70&-&-&-&17.30&11.00&14.30\\ 
				&RCIL*~\cite{RCIL}&CVPR2022&DeepLabv3&75.00&42.80&67.30&66.10&18.20&54.70 &-&-&-&30.60&4.70&18.20 \\
				&Incrementer*~\cite{Incrementer}&CVPR2023&ViT-B/16&81.59&62.17&77.60&81.42&57.05&76.25&-&-&-&77.62&60.33&70.16 \\
				\midrule
				\multicolumn{15}{c}{Overlapped} \\
				\midrule
				\multirow{21}*{\begin{sideways}Data-free\end{sideways}} &\emph{fine tuning}&-&DeepLabv3&2.10&33.10&9.80&0.20&1.80&0.60&0.50&10.40&7.60&6.30&2.80&4.70\\
				&EWC*~\cite{EWC} &PNAS2017&DeepLabv3&24.30 &35.50 &27.10 &0.30 &4.30 &1.30 &-&-&-&- &- &-\\
				&LwF-MC*~\cite{iCaRL}&CVPR2017&DeepLabv3 &58.10&35.00&52.30&6.40&8.40&6.90&20.91&36.67&24.66&4.65&5.90&4.95\\
				&ILT*~\cite{ILT} &ICCVW2019&DeepLabv3&66.30 &40.60&59.90&4.90&7.80&5.70&22.51&31.66&29.04&7.15&3.67&5.50\\
				&MiB*~\cite{MiB}&CVPR2020&DeepLabv3&76.37&49.97&70.08&34.22&13.50&29.29&57.10&42.56&46.71&12.25&13.09&12.65\\
				&SSUL*~\cite{SSUL}&NeurIPS2021&DeepLabv3&77.42&47.16&70.21&78.06&28.54&66.27&71.17&45.38&52.75&73.78&41.13&58.23\\
				&PLOP*~\cite{PLOP}&CVPR2021&DeepLabv3&75.73&51.71&70.09&65.12&21.11&54.64&17.48&19.16&18.68&44.03&15.51&30.45 \\
				&UCD+PLOP~\cite{UCD}&TPAMI2022&DeepLabv3&75.00&51.80&69.20&66.30&21.60&55.10&-&-&-&42.30&28.30&35.30\\
				&REMINDER*~\cite{REMINDER} &CVPR2022&DeepLabv3&76.11&50.74&70.07&68.30&27.23&58.52&-&-&-&-&-&- \\
				&RCIL*~\cite{RCIL}&CVPR2022&DeepLabv3&78.80&52.00&72.40&70.60&23.70&59.40&65.30&41.49&50.27&55.40&15.10&34.30\\
				&RBC*~\cite{RBC}&ECCV2022&DeepLabv3&76.59&52.78&70.92&69.54&38.44&62.14&-&-&-&-&-&-\\
				&SPPA*~\cite{SPPA}&ECCV2022&DeepLabv3&78.10&52.90&72.10&66.20&23.30&56.00&-&-&-&-&-&-\\
				&CAF*~\cite{CAF}&TMM2022&DeepLabv3&77.20&49.90&70.40&55.70&14.10&45.30&-&-&-&-&-&-\\
				&DKD*~\cite{DKD} &NeurIPS2022&DeepLabv3&78.83&58.23&73.93&78.09&42.72&69.67&-&-&-&-&-&- \\
				&SATS*~\cite{SATS}&PR2023&SegFormerB2&80.24&61.17&75.70&78.38&62.02&74.48&75.43&64.13&67.36&64.27&58.66&61.60 \\
				&AWT+MiB*~\cite{AWT}&WACV2023&DeepLabv3&77.30&52.90&71.50 &59.10&17.20&49.10&61.80&45.90&50.40&33.20&18.00&26.00\\
				&EWF+MiB*~\cite{EWF}&CVPR2023&DeepLabv3&-&-&-&78.00&25.50&65.50&69.00&45.00&51.80&56.00&16.70&37.30\\
				&IDEC~\cite{IDEC}&TPAMI2023&DeepLabv3&78.01&51.84&71.78&76.96&36.48&67.32&67.05&48.98&54.14&70.74&46.30&59.10\\
				&FMWISS*~\cite{FMWISS}&CVPR2023&DeepLabv3&78.40 & 54.50& 73.30&-&-&-&-&-&-&- &-&- \\
				&Incrementer*~\cite{Incrementer}&CVPR2023&ViT-B/16&82.53&69.25&79.93& 79.60&59.56&75.55&-&-&-&77.62&60.33&70.16\\
				&GSC*~\cite{GSC}&TMM2024&DeepLabv3&78.30&54.20&72.60&72.10&24.40&60.80&-&-&-&50.60&17.30&34.70 \\
				&CoMasTRe*~\cite{CoMasTRe}&CVPR2024&Mask2Former&79.73&51.93&73.11&69.77&43.62&63.54&-&-&-&-&-&-\\
				\midrule
				\multirow{11}*{\begin{sideways}Data-replay\end{sideways}} 
				&SDR*~\cite{SDR} &CVPR2021&DeepLabv3+&75.40&52.60&69.90&44.70&21.80&39.20&-&-&-&32.40&17.10&25.10 \\
				&RECALL-GAN~\cite{RECALL}&ICCV2021&DeepLabv2&66.60&50.90&64.00&65.70&47.80&62.70&-&-&-&59.50&46.70&54.80\\
				&RECALL-Web~\cite{RECALL}&ICCV2021&DeepLabv2&67.70&54.30&65.60&67.80&50.90&64.80&-&-&-&65.00&53.70&60.70\\
				&SSUL-M*~\cite{SSUL}&NeurIPS2021&DeepLabv3&79.53&52.87&73.19&78.92&43.86&70.58&72.97&49.02&55.85&74.79&48.87&65.45\\
				&SPPA*~\cite{SPPA}&ECCV2022&DeepLabv3&78.10&52.90&72.10&66.20&23.30&56.00&-&-&-&-&-&-\\
				&MicroSeg-M*~\cite{MicroSeg}&NeurIPS2022&DeepLabv3&82.00&59.20&76.60&81.30&52.50&74.40&74.80&60.50&64.60&77.20&57.20&67.70\\
				&DKD-M*~\cite{DKD} &NeurIPS2022&DeepLabv3&79.13&60.59&74.72&78.84&52.36&72.53&-&-&-&-&-&- \\
				&SATS-M*~\cite{SATS}&PR2023&SegFormerB2&81.44&70.02&78.72&80.37&64.54&76.61&75.58&69.67&71.36&76.21&61.62&69.27\\
				&AMSS*~\cite{AMSS}&CVPR2023&DeepLabv3&79.31&55.88&73.73& 78.54&50.82&71.94&-&-&-&-&-&- \\
				&TIKP*~\cite{TIKP}&AAAI2024&DeepLabv3&78.81&55.50&73.26&73.77&42.31&66.28&-&-&-&69.71&43.48&57.22 \\
				&LAG*~\cite{LAG}&TPAMI2024&DeepLabv3&77.33&51.76&71.24&75.00&37.52&66.08&67.53&47.11&52.94&69.56&42.62&56.73\\
				\midrule
				&\emph{offline} &-&DeepLabv3&79.77&72.35&77.43&79.77&72.35&77.43&76.91&77.63&77.43&78.41&76.35&77.43\\
				&\emph{offline} &-&SegFormerB2&80.84&74.97&79.44&80.84&74.97&79.44&78.36&79.87&79.44&80.46&78.32&79.44 \\
				\bottomrule[0.5mm]
		\end{tabular*}}{}}	
	\label{table-VOC2012}
\end{table*}

In Table~\ref{table-VOC2012}, we report the IoU performance on the old and new classes respectively to reveal the anti-forgetting performance and new-knowledge learning performance. Additionally, the overall performance after all CL steps is also calculated as a balance of plasticity and stability. We also report two baselines for reference, i.e., \emph{fine-tuning} on $C^{t}$ and training on all classes \emph{offline}. The former is the lower bound and the latter can be regarded as the upper bound in CSS tasks. 

1) \textbf{\textit{Dependence on old data}}: In general,  replay-based methods achieve higher IoU in both old classes and new classes than data-free methods. For example, SSUL-M~\cite{SSUL}  introduces exemplar-memory to achieve 65.45\% mIoU of all classes on VOC 10-1, which exceeds SSUL (58.23\%) with a 7.22\% margin. 

2) \textbf{\textit{Efficiency on incremental data}}: Currently many CSS methods propose to alleviate the burden of labeled incremental data. They focus on few-/zero-shot learning manner or weakly-supervised manner. For example, FMWISS~\cite{FMWISS} introduces large-model-based co-segmentation to generate dense masks based on image-level labels to achieve weakly-supervised CSS. It also achieve remarkable performance compared with fully-supervised methods. LAG~\cite{LAG} explores class-incremental CSS under limited incremental data and achieves favourable performance.

3) \textbf{\textit{Effectiveness of knowledge distillation}}:  As an indispensable manner in CSS, KD is tasked with inheriting knowledge from the old model. ILT~\cite{ILT} and MiB~\cite{MiB} anticipatorily utilize KD in intermediate representations and output logits, which bring a prospect on MSFC tasks. Further PLOP~\cite{PLOP} and IDEC~\cite{IDEC} propose multi-level distillation strategies to boost CSS performance. For example, PLOP achieves 30.45\% mIoU on VOC 10-1 task, which proves the effectiveness of multi-level KD compared to MiB (12.65\%). Current up-to-date methods usually introduce additional regularization terms based on KD. For example, IDEC proposes an asymmetric region-wise contrastive learning manner aligning with multi-level KD to achieve 59.10\% mIoU on MSFC VOC 10-1 task.  

4) \textbf{\textit{Impact of segmentation model}}: For a fair comparison, many CSS methods directly employ an existing semantic segmentation model such as DeepLabv3~\cite{DeepLabv3} with pretrained backbone. To reveal the impact of different segmentation models and backbones,  IDEC~\cite{IDEC} and LAG~\cite{LAG} proceeds with an ablation study including two semantic segmentation models with CNN and Transformers as backbones. Similarly, SATS~\cite{SATS} uses SegFormer~\cite{SegFormer} as segmentation model and achieves 61.60\% mIoU on VOC 10-1. Incrementer~\cite{Incrementer} reports 70.16\% mIoU on VOC 10-1 with ViT~\cite{ViT}. The quantitative results from~\cite{IDEC, LAG, Incrementer, SATS} prove that stronger segmentation models can achieve superior CSS performance on both old and new classes.

\noindent\textbf{ADE20K}. On ADE20K,  we select four representative settings 100-50 (2 steps), 100-10 (6 steps), 50-50 (3 steps) and 100-5 (11 steps). Among these settings, 100-50 and 50-50 are FSMC means, 100-5 is MSFC setting and 100-10 is MSMC manner. All results are based on the \textit{overlapped} setting since it is more realistic and challenging.

\begin{table*}[htbp]
	\centering
	\scriptsize
	\caption{Class-incremental CSS quantitative comparison on ADE20K in mIoU (\%) under \textit{overlapped} setting. Methods with * indicate the results were directly taken from the corresponding original work.}
	\setlength{\tabcolsep}{1mm}{
		{\begin{tabular*}{0.95\textwidth}{@{\extracolsep{\fill}}ll|c|c|ccc|ccc|ccc|ccc@{}}
				\toprule[0.5mm]
				&\multirow{2}*{Method}&\multirow{2}*{Year} &\multirow{2}*{Model}&
				\multicolumn{3}{c}{100-50 (2 steps)} & \multicolumn{3}{c}{100-10 (6 steps)} &\multicolumn{3}{c}{50-50 (3 steps)} & \multicolumn{3}{c}{100-5 (11 steps)}\\
				&&&& 1-100 & 101-150 & all & 1-100 & 101-150 & all & 1-50 & 51-150 & all  & 1-100 & 101-150 & all\\
				\midrule
				\multirow{17}*{\begin{sideways}Data-free\end{sideways}}&\emph{fine tuning}&-&DeepLabv3&0.00&11.22&3.74&0.00&2.08&0.69&0.00&3.60&2.40&0.00&0.07&0.02\\
				&ILT*~\cite{ILT} &ICCVW&DeepLabv3&18.29&14.40&17.00&0.11&3.06&1.09 &3.53&12.85&9.70&0.08&1.31&0.49\\
				&MiB*~\cite{MiB}&CVPR2020&DeepLabv3&40.52&17.17&32.79&38.21&11.12&29.24&45.57&21.01&29.31&36.01&5.66&25.96\\
				&SSUL*~\cite{SSUL}&NeurIPS2021&DeepLabv3&- &-&-&42.10&16.02&33.46&-&-&-&42.03&15.80&33.35\\
				&PLOP*~\cite{PLOP}&CVPR2021&DeepLabv3&41.87&14.89&32.94&40.48&13.61&31.59&48.83&20.99&30.40&39.11&7.81&28.75\\
				&UCD+PLOP~\cite{UCD}&TPAMI2022&DeepLabv3&42.12&15.84&33.31&40.80&15.23&32.29&47.12&24.12&31.79&-&-&-\\
				&REMINDER*~\cite{REMINDER}&CVPR2022&DeepLabv3 &41.55&19.16&34.14&38.96&21.28&33.11&47.11&20.35&29.39&36.06&16.38&29.54 \\
				&RCIL*~\cite{RCIL}&CVPR2022&DeepLabv3&42.30&18.80&34.50&39.30&17.60&32.10&48.30&25.00&32.50&38.50&11.50&29.60\\
				&SPPA*~\cite{SPPA}&ECCV2022&DeepLabv3&42.90&19.90&35.20&41.00&12.50&31.50&49.80&23.90&32.50&-&-&-\\
				&DKD*~\cite{DKD} &NeurIPS2022&DeepLabv3&42.41&22.89&35.95&41.56&19.51&34.26&48.84&26.28&33.90&-&-&- \\
				&SATS*~\cite{SATS}&PR2023&SegFormerB2&-&-&-&41.42&19.09&34.18&-&-&-&-&-&- \\
				&AWT+MiB*~\cite{AWT}&WACV2023&DeepLabv3&40.90&24.70&35.60&39.10&21.40&33.20&46.60&27.00&33.50&38.60&16.00&31.10\\
				&EWF+MiB*~\cite{EWF}&CVPR2023&DeepLabv3&41.20&21.30&34.60&41.50&16.60&33.20&-&-&-&41.40&13.40&32.10\\
				&IDEC~\cite{IDEC}&TPAMI2023&DeepLabv3&42.01&18.22&34.08&40.25&17.62&32.71&47.42&25.96&33.11&39.23&14.55&31.00\\
				&Incrementer*~\cite{Incrementer}&CVPR2023&ViT-B/16&49.42&35.62&44.82&48.47&34.62&43.85&56.15&37.81&43.92&46.93&31.31&41.72\\
				&GSC*~\cite{GSC}&TMM2024&DeepLabv3&42.40&19.20&34.80&40.80&16.20&32.60& 46.20 &26.40&33.00&-&-&-\\
				&CoMasTRe*~\cite{CoMasTRe}&CVPR2024&Mask2Former&45.73&26.02&39.20&42.32&18.42&34.41&-&-&-&40.82&15.83&32.55\\
				\midrule
				\multirow{7}*{\begin{sideways}Data-replay\end{sideways}} 
				&SSUL-M*~\cite{SSUL}&NeurIPS2021&DeepLabv3&42.20&13.95&32.80&42.17&16.03&33.89&49.55&25.89&33.78&42.53&15.85&34.00\\
				&SPPA*~\cite{SPPA}&ECCV2022&DeepLabv3&42.90&19.90&35.20&41.00&12.50&31.50&49.80&23.90&32.50&-&-&-\\
				&MicroSeg-M*~\cite{MicroSeg}&NeurIPS2022&DeepLabv3&43.40&20.90&35.90&43.70&22.20&36.60&49.80&22.00&31.40&43.60&22.40&36.60\\
				&DKD-M*~\cite{DKD}&NeurIPS2022&DeepLabv3&42.43&22.95&35.98&41.74&20.11&34.58&48.84&26.31&33.92&-&-&- \\
				&AMSS*~\cite{AMSS}&CVPR2023&DeepLabv3&44.06&24.96&37.74&43.88&25.14&37.67&-&-&-&43.35&18.53&35.13 \\
				&TIKP*~\cite{TIKP}&CVPR2024&DeepLabv3&42.17&20.21&34.90&40.96&19.56&33.79&48.75&25.86&33.56&37.48&17.56&30.88\\
				&LAG*~\cite{LAG}&TPAMI2024&DeepLabv3&41.64&19.73&34.34&41.00&18.69&33.56&47.69&26.12&33.31&39.96&17.22&32.38\\
				\midrule
				&\emph{offline}&-&DeepLabv3 &44.30&28.20&38.90&44.30&28.20&38.90&50.90&32.90&38.90&44.30&28.20&38.90\\
				&\emph{offline}&-&ViT-B/16&49.79&37.09&45.56&49.79&37.09&45.56&56.43&40.12&45.56&49.79&37.09&45.56\\
				\bottomrule[0.5mm]
		\end{tabular*}}{}}	
	\label{table-ADE20K}
\end{table*}

As seen in Table~\ref{table-ADE20K}, compared with VOC 2012, ADE20K is more challenging due to the large number of classes and the complex semantics distribution. In 100-10 task, KD-based methods~\cite{MiB, PLOP} encounter severe semantic drift on new classes reflected by low IoU. Considering the upper bound mIoU is only 38.9\% (DeepLabv3), it suggests significant pixel misclassification. However, a stronger segmentation model may bring more evident improvement in balancing plasticity and stability~\cite{Incrementer}. Thus we propose a hypothesis: \textit{\textbf{How to evaluate CSS performance objectively but with a certain emphasis?}}
We discuss this problem from two aspects:  1) For easy CSS tasks like VOC 15-5, the primary focuses should be on the CL strategies. It is because the anti-forgetting of the old classes can be guaranteed by the model itself, it is necessary to focus on learning the new class and suppressing semantic drift. 2) For hard CSS tasks like ADE 100-5, effort should be put into increasing the performance of semantic segmentation models. The reason is the severe catastrophic forgetting aggravated by the limited performance of the segmentation model.

\subsubsection{Domain-incremental CSS Evaluation}
Domain-incremental CSS focuses on exploring how to teach a model to recognize semantics in images across different domains. The model is incrementally updated by adapting its segmentation capabilities to new domains. Typically, the semantic classes in domain-incremental CSS remain unchanged. Here we would like to discuss the relation and difference between \textit{domain adaptive semantic segmentation} (DASS) and \textit{domain-incremental CSS} (DICSS). Both of them transfer a model from one domain to other unseen domains for the model's continual updating. The main difference lies in the task objective. Concretely, DASS only highlight the performance on the new domains, while DICSS considers both the old and the new domains to achieve proper compatibility between stability and plasticity. 

\begin{table}[tbp]
	\centering
	\footnotesize
	\caption{Class-incremental CSS quantitative comparison on Cityscapes in mIoU (\%). Methods with * indicate the results were directly taken from the original work. Methods with $\dagger$ mean the results are from~\cite{RCIL}. }
	\setlength{\tabcolsep}{1mm}{
		{\begin{tabular*}{0.42\textwidth}{@{\extracolsep{\fill}}l|ccc@{}}
				\toprule[0.5mm]
				Method&11-5 (3 steps)& 11-1 (11 steps) & 1-1 (21 steps)\\
				\midrule
				\textit{fine-tuning}$^{\dagger}$&61.7&60.4&42.9\\
				LwF$^{\dagger}$~\cite{LWF}&59.7&57.3&33.0\\
				LwF-MC$^{\dagger}$~\cite{LWF}&58.7&57.0&31.4\\
				ILT$^{\dagger}$~\cite{ILT} &59.1&57.8&30.1\\
				MiB$^{\dagger}$~\cite{MiB} & 61.5&60.0&42.2\\
				PLOP*~\cite{PLOP} &63.5&62.1&45.2\\
				RCIL*~\cite{RCIL} &64.3&63.0&48.9\\
				\bottomrule[0.5mm]
		\end{tabular*}}{}}	
	\label{table-Cityscapes}
\end{table}
\noindent\textbf{Cityscapes}. Taking Cityscapes~\cite{Cityscapes} as a benchmark, we investigate current representative DICSS methods on 11-5 (3 steps), 11-1 (11 steps) and 1-1 (21 steps) in Table~\ref{table-Cityscapes}. The key evaluation focuses on the average accuracy across all domains after all CL steps. It is noticeable that \textit{fine-tuning} manner achieves favourable performance compared with other CSS methods, which is because the different domains across Cityscapes possess small domain gap in appearance and semantics. 

\begin{figure*}[htbp]
	\centering
	\begin{minipage}[t]{0.48\linewidth}
		\centering
		\includegraphics[scale=0.4]{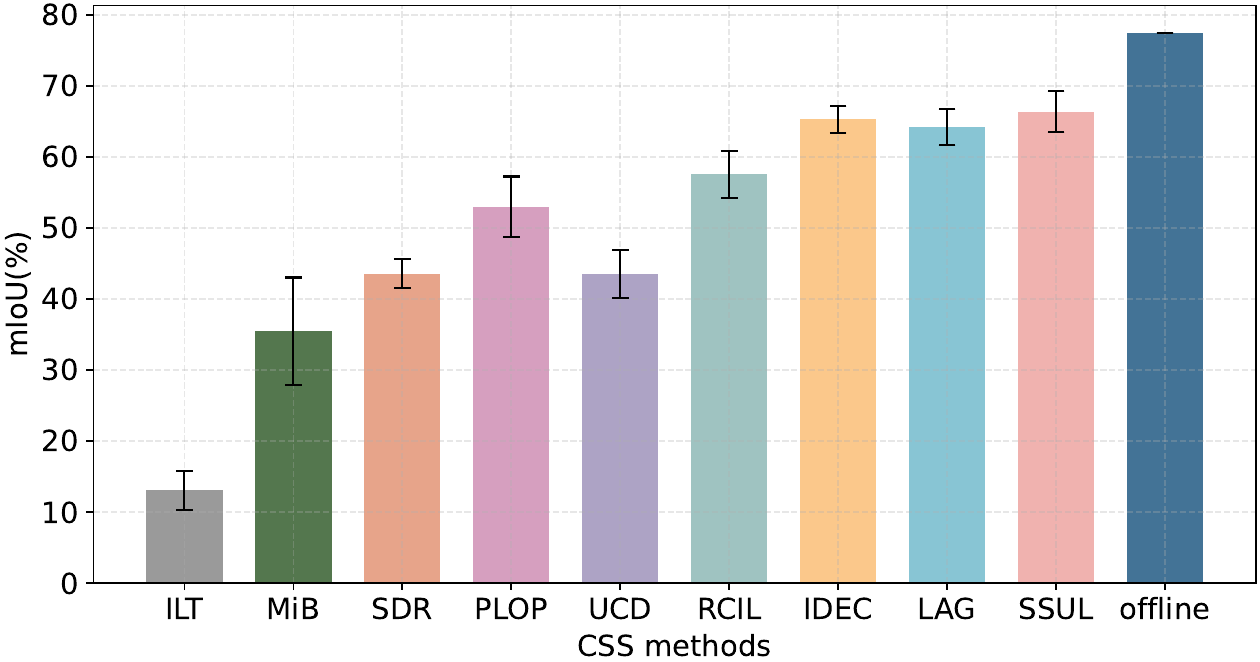}
		\caption{The average performance and standard deviation under various incremental class orders on VOC 15-1 task under \textit{overlapped} setting.}
		\label{fig-mIoU_class_orders}
	\end{minipage}	
	\begin{minipage}[t]{0.48\linewidth}
		\centering
		\includegraphics[scale=0.4]{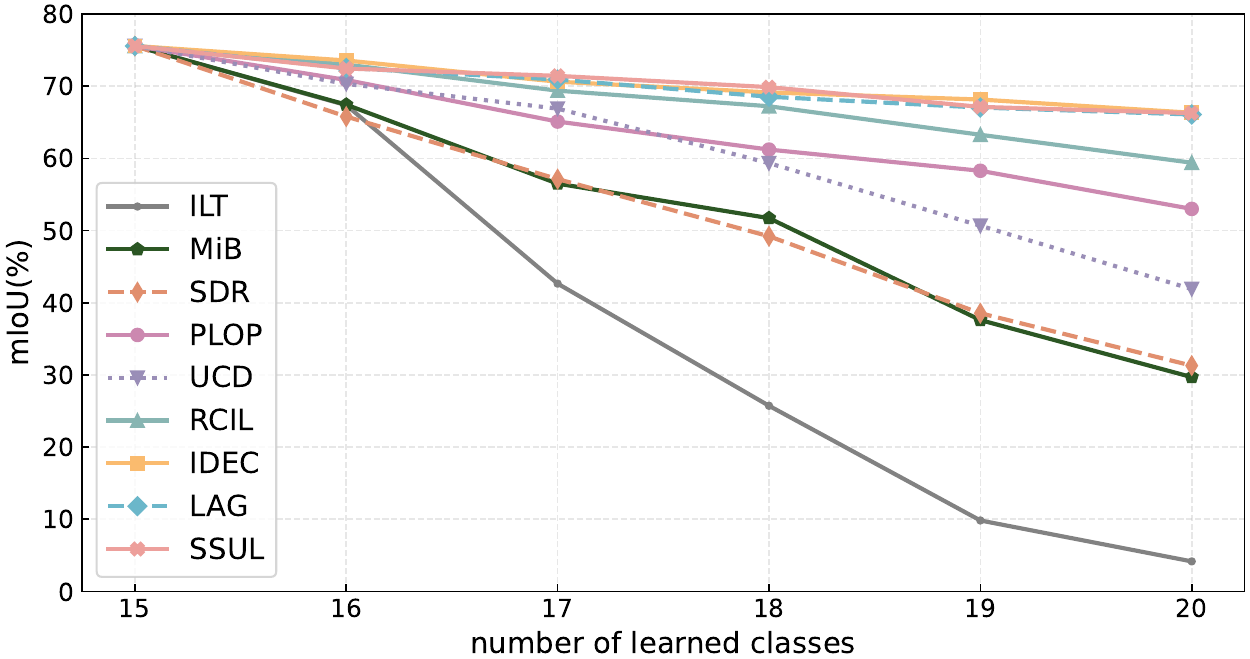}
		\caption{The mIoU (\%) evolution against number of learned classes on VOC 15-1 task under \textit{overlapped} setting.}
		\label{fig-mIoU_curve}
	\end{minipage}
\end{figure*}	
\subsubsection{Robustness Analysis}
In CSS, model robustness is reflected in anti-forgetting on learned knowledge and various CL settings. Thus the robustness of CSS models can be quantitatively evaluated via class incremental orders and performance after CL steps. 
\noindent\textbf{Robustness to Class Incremental Orders}.
We perform class-incremental CSS experiments on VOC 15-1 with five different class orders including the ascending order and four random orders as follows.
\begin{equation}
	\scriptsize
	\begin{split}
		\nonumber
		a:\{[0,1,2,3,4,5,6,7,8,9,10,11,12,13,14,15],[16],[17],[18],[19],[20]\} \\
		b:\{[0,12,9,20,7,15,8,14,16,5,19,4,1,13,2,11],[17],[3],[6],[18],[10]\} \\
		c:\{[0,13,19,15,17,9,8,5,20,4,3,10,11,18,16,7],[12],[14],[6],[1],[2]\} \\
		d:\{[0,15,3,2,12,14,18,20,16,11,1,19,8,10,7,17],[6],[5],[13],[9],[4]\} \\
		e:\{[0,7,5,3,9,13,12,14,19,10,2,1,4,16,8,17],[15],[18],[6],[11],[20]\} \\
	\end{split}
\end{equation}

As seen in Fig.~\ref{fig-mIoU_class_orders}, the average mIoU performance with a standard deviation of several representative CSS methods~\cite{ILT, MiB, SDR, PLOP, RCIL, IDEC, LAG, SSUL} is reported. The higher mIoU and more limited deviation indicate the model achieves better balance between plasticity and stability. the data-replay method SSUL achieves superior performance to the other up-to-date data-free method.

\begin{figure}[tbp]
	\centering
	\includegraphics[scale=0.52]{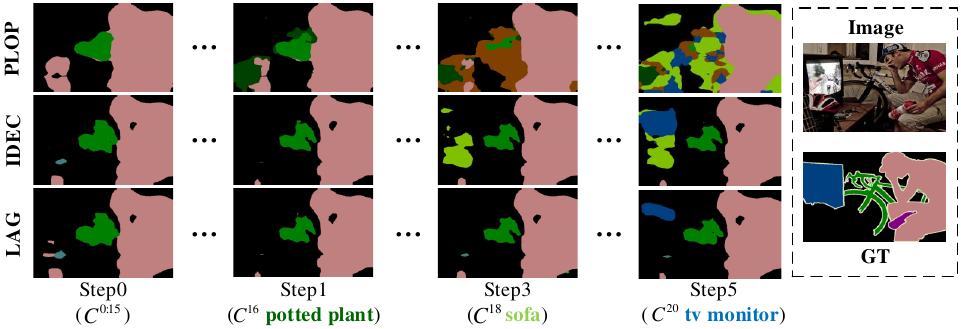}
	\caption{Qualitative results of various CSS approaches~\cite{PLOP, IDEC, LAG} on VOC 15-1 task.}
	\label{fig-qualitative_visualization}
\end{figure}

\noindent\textbf{Robustness to CL Steps}.
In CSS tasks, catastrophic forgetting occurs during the continuous updating process. Therefore, a valid metric that measures the anti-forgetting ability of CSS models is reflected in the model's performance on both new and old data after CL steps.  As shown in Fig.~\ref{fig-mIoU_curve}, we evaluate mIoU on all classes against the number of learned classes on VOC 15-1 under overlapped setting in terms of current up-to-date CSS methods~\cite{ILT, MiB, SDR, PLOP, RCIL, IDEC, LAG, SSUL}. For example, ILT experiences severe forgetting, evident in the rapid decline in model performance with increase in CL steps. In contrast, SSUL maintains a higher resistance to forgetting while ensuring the ability to learn new classes, as reflected in the overall performance decline not being significant after all CL steps. Besides, the qualitative visualizations of several CSS methods are shown in Fig.~\ref{fig-qualitative_visualization}. With new semantics continuously arriving, the forgetting and semantic drift problems are reflected by the pixel misclassifications.

\subsubsection{Interpretability Analysis}
Model interpretability assists the analysis of semantic changes, feature distributions and the possibility of revealing forgetting in the CL process. Effective manners include feature-based visualization~\cite{TSNE, GradCAM}, layer-wise relevance propagation~\cite{LRP}, similarity in representations~\cite{ramasesh2020anatomy, kalb2022causes}, linear probing~\cite{davari2022probing} and linear mode connectivity~\cite{mirzadeh2020linear}, etc. In this section, we apply the TSNE visualization for intuitively explaining the model changes before and after CL steps.

\begin{figure}[htbp]
	\centering
	\includegraphics[scale=0.25]{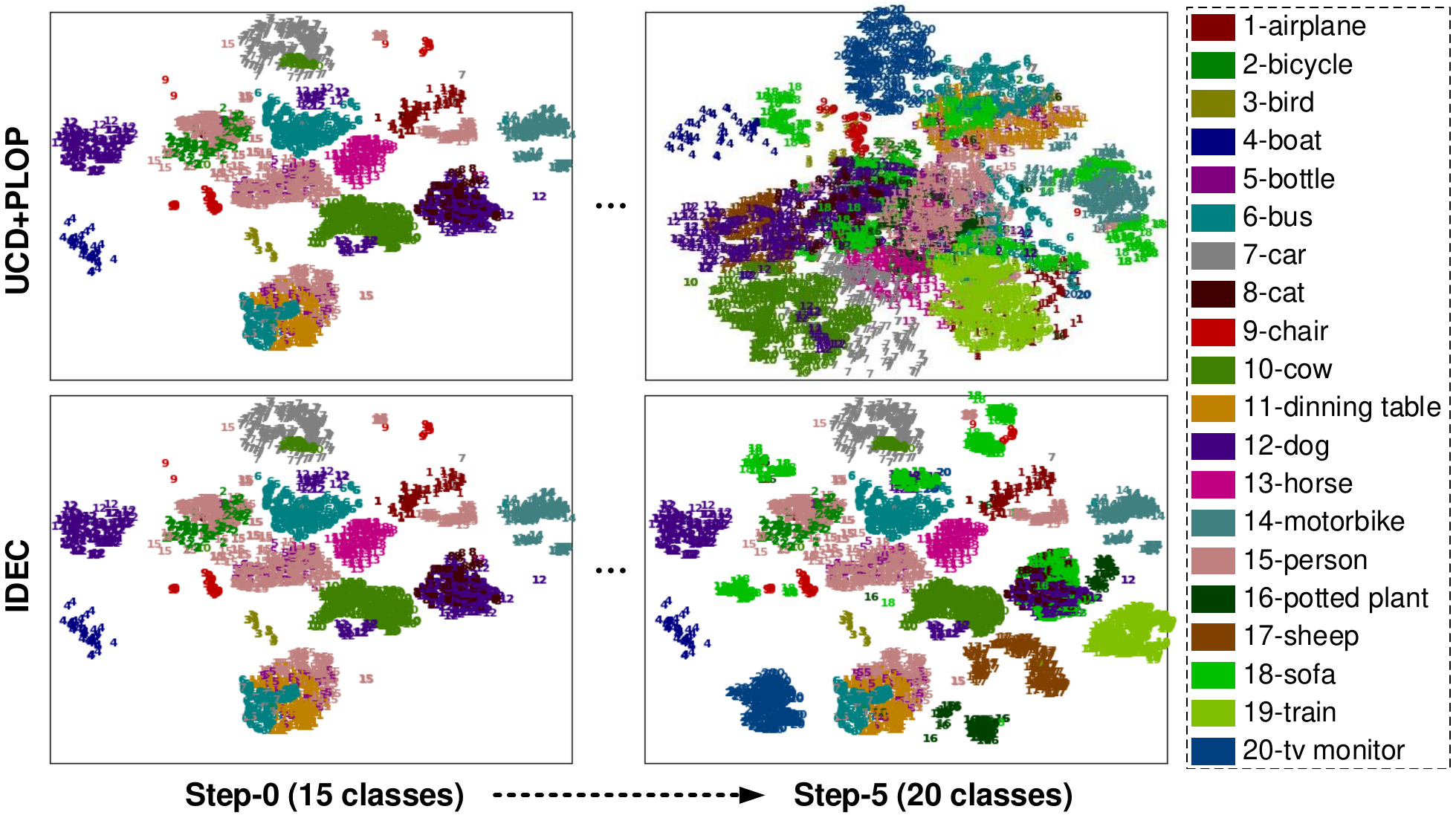}
	\caption{TSNE~\cite{TSNE} map generated from~\cite{IDEC} for class-incremental VOC 15-1 task. The number in the image represents the corresponding class. It intuitively shows the feature distribution before and after the CL steps. The background class is ignored for clearer visualization.}
	\label{fig-tsne_IDEC}
\end{figure}
Seeking to the continuous adaptation to newly added data or semantics, CSS models require constant adjustments of their parameters. Therefore, analyzing the changes within the model is a prerequisite for interpreting CL process. Explainability analysis can assist in comprehending how the model adapts to new data, thereby enhancing the reliability in the model. For example, the class clusters vary in class-incremental CSS scenarios. Thus visualizing the feature distribution in high-dimension feature space can disclose the core reason of catastrophic forgetting and reveal semantic variance. T-SNE~\cite{TSNE} maps the high-dimensional features to low-dimensional space, which is suitable for investigating the inner feature distribution after incremental steps. As seen in Fig.~\ref{fig-tsne_IDEC}, we present the TSNE visualizations of two representative CSS approaches including UCD~\cite{UCD}+PLOP~\cite{PLOP} and IDEC~\cite{IDEC} on VOC 15-1 task at the initial step and the final step, respectively. On the one hand, the TSNE map intuitively shows the catastrophic forgetting, which is reflected by the shift cluster center of the initially-learned classes after CL steps. On the other hand, it also reveals the IL ability since the incremental classes are clustered into new clusters in the feature space. Other interpretability tools like LRP~\cite{LRP}, which is explored in~\cite{LAG}, are also validate and helpful for improving the interpretability of CSS models.

\section{Applications and Prospects}
\label{Sec-Applications}
\subsection{Applications}
\textbf{Autonomous driving}: Class-\&domain-incremental CSS methods allow the model to learn new classes and new domains over time. This is crucial in autonomous driving scenarios where new objects or road conditions may emerge. Techniques like knowledge distillation and feature replay are explored to facilitate CSS in autonomous driving systems. For example, Barbato et al.~\cite{ContinualPMF} propose a modality-incremental manner for multi-modal 3D semantic segmentation, which processes LiDAR and RGB data for road-scene semantic segmentation. Kalb et al.~\cite{Kalb_2023_CVPR} explore the causes of catastrophic forgetting in adverse weather conditions for domain-incremental CSS. Additionally, considering the joint interpretation of multi-modal data such as RGB, LiDAR, etc., CSS models need to address challenges related to unsupervised domain-incremental adaptation~\cite{Truong2022CONDACU}, multi-modal data alignment~\cite{MM-CTTA} and multi-task learning~\cite{liang2022effective}. 

\noindent\textbf{In-orbit remote-sensing observation}: Remote sensing satellites continuously provide a vast amount of time-series incremental data, such as land cover changes and meteorological observations. In this field, CSS can assist the in-orbit system in monitoring and analyzing these data self-intelligently under constantly arriving data conditions~\cite{rong2022historical, marsocci2023continual, rui2023dilrs, MiSSNet}, including atmospheric pollution, soil quality, forest health, change detection~\cite{MDINet}, etc. When new monitoring requirements or tasks emerge, the system can adjust its monitoring methods adaptively. Considering the constraints on in-orbit observation computing and storage resources, in-orbit CSS model deploying and self-evolving under the conditions of edge computing and limited data storage will also become a research focus.

\noindent\textbf{Auxiliary medical diagnosis}: In the context of automated lesion tracking and monitoring, CSS can provide more accurate image analysis, earlier disease detection, personalized medical care, and more efficient medical practices. For instance, it can be used to discern newly added lesion locations or disease types~\cite{Ji_2023_ICCV, liu2022learning}, generate customized diagnoses and treatment plans based on a patient's specific condition, which is crucial for improving patient survival rates and treatment effectiveness. However, in medical imaging, one of the most crucial performance aspects is achieving the most accurate diagnoses. Therefore, the requirements for a model's anti-forgetting capacity and its ability to learn new knowledge are exceptionally stringent. The current dilemma lies in the fact that maintaining separate models leads to increased computational resource costs while retaining a unified model faces challenges related to accuracy and inherent privacy risks~\cite{gonzalez2020wrong}.

\subsection{Future Prospects}
After nearly a decade of development, CSS has gained much more attention not only in theoretical exploration but also in task extension and application. However, when facing the real-world application, research on CSS still has a long way ahead from algorithms to applications. While there are many difficulties and challenges, it is encouraging that CSS has already demonstrated significant application value and development prospects. We offer the following perspectives on technical challenges and future trends in CSS:
\begin{itemize}
\item [1)] \textit{Brain-like Modeling}: The human brain is capable of accumulating new knowledge, rapidly processing multi-modal information, and exhibiting highly knowledge-association ability with low energy consumption. 
Research on CSS models based on brain-like mechanisms holds promise for addressing catastrophic forgetting and achieving solid knowledge accumulation.
\item [2)] \textit{Interpretability Modeling}: Extending explainability of continual learning settings, which is crucial for understanding model updates and adaptation and improving model trustworthiness.
\item [3)] \textit{Human-AI Collaboration}: Exploring CSS approaches that facilitate collaboration between AI models and human experts, allowing users to provide feedback and corrections to improve the model's application in embodied AI systems.
\item [4)] \textit{Cross-modality Incremental Adaptation}:  Modality-incremental learning across multi-domain and multi-task has a strong application prospect in open-world understanding. The technical challenge lies in achieving compatibility and coexistence of new and old knowledge under substantial task variation and significant differences of multi-modal data.
\item [5)] \textit{Online and Active Learning}: Online learning allows CSS models to continuously acquire data from real-world systems and continuously self-evolving. Active learning techniques can assist in selecting the most informative data for continual learning.
\item [6)] \textit{Hardware Acceleration and Edge Computing}: To cater to embedded devices and edge computing applications such as autonomous driving and in-orbit intelligent interpretation, future CSS methods will require efficient hardware acceleration and model compression techniques to meet real-time and resource-constrained application.
\end{itemize}

\section{Conclusion}
Continual semantic segmentation (CSS) enables a model to continuously learn new knowledge while maintains retention of existing knowledge in dynamic and open environments, striking a balance between stability and plasticity. This technique closely mimics human learning mechanisms and holds significant value for building strong artificial intelligence, expanding its application domains, and enhancing its service levels in human life.

Over the past decade, CSS has been witnessed its origin, development and flourishing. In this paper, we are committed to introducing a valuable survey on CSS.  We present a comprehensive review of problem definitions, challenges, methodologies, cutting-edge advancements, qualitative and quantitative analysis, and diverse applications of this expertise field. We categorize CSS into two routes including five sub-categories and four specialties, covering the comprehensive research in the field.
Research in this area spans many intersecting fields including biology, neuroscience, artificial neural networks, computer vision, etc. Consequently, CSS has yielded a large number of research achievements.  This review is designed not only to benefit researchers in the field but also to facilitate interdisciplinary collaboration and engagement from researchers in various domains.  Future CSS studies will concentrate on exploring the coupling between human cognition patterns and machine learning models. We believe that CSS models will evolve towards greater intelligence, robustness, interpretability and wider application prospects.

\ifCLASSOPTIONcaptionsoff
  \newpage
\fi

{\small
	\bibliographystyle{IEEEtran}
	\bibliography{refs}
}

\end{document}